\documentclass[journal]{IEEEtai}

\usepackage[colorlinks,urlcolor=blue,linkcolor=blue,citecolor=blue]{hyperref}

\usepackage{color,array}

\usepackage{graphicx}

\usepackage[cmex10]{amsmath}
\usepackage{array}
\usepackage{mdwmath}
\usepackage{mdwtab}
\usepackage{eqparbox}
\usepackage{url}
\usepackage{multirow}

\usepackage{amsmath}
\usepackage{booktabs}
\usepackage{algorithm}
\begin{document}

\title{Irony Detection, Reasoning and Understanding in Zero-shot Learning}

 \author{
     \IEEEauthorblockN{Peiling Yi, Yuhan Xia, and Yunfei Long}
 }

\maketitle

\begin{abstract}
Irony is a powerful figurative language (FL) on
social media that can potentially mislead various NLP tasks,
such as recommendation systems, misinformation checks, and
sentiment analysis. Understanding the implicit meaning of this
kind of subtle language is an essential step to mitigate the negative
impact of irony in NLP tasks. However, existing efforts are limited to domain-specific datasets and struggle to generalize across diverse real-world scenarios. Moreover, reasoning for model decisions that accurately capture semantic and affective meaning remains underexplored. To address these limitations, this paper proposes a conceptual framework called IDADP, which leverages Large language models(LLMs)' in-context learning capabilities to detect irony and generate human-like explanations across diverse datasets and platforms without prior training on ironic samples. Extensive experiments on six widely used irony detection datasets, utilising two large language models (GPT and Gemini), demonstrate that IDADP consistently outperforms six competitive zero-shot baselines and approaches the performance of three fine-tuned supervised learning baselines. Additionally, we examine GPT's ability to understand the true intent behind ironic text within the IDADP framework, highlighting its strong potential to recognize and interpret statements where the intended meaning differs from or contrasts with the literal meaning. Furthermore, we conduct qualitative analyses to identify remaining challenges. This work, in turn, opens an avenue for transparent decision-making in irony detection. 
\end{abstract}

\begin{IEEEImpStatement}

The generalization of irony detection faces significant challenges that lead to substantial performance deviations when detection models are applied to diverse real-world scenarios. In the study, we find that irony-focused prompts, as generated from our IDADP framework for LLMs, can not only overcome dataset-specific limitations but also generate coherent, human-readable reasoning, transforming ironic text into its intended meaning. Based on our findings and in-depth analysis, we identify several promising directions for future research aimed at enhancing LLMs’ zero-shot capabilities in irony detection, reasoning, and comprehension. These include advancing contextual awareness in irony detection, exploring hybrid symbolic-neural methods, and integrating multimodal data, among others.

\end{IEEEImpStatement}


\section{Introduction}
\begin{figure*}[tbh]
  \centering
  \includegraphics[width=\linewidth]
  {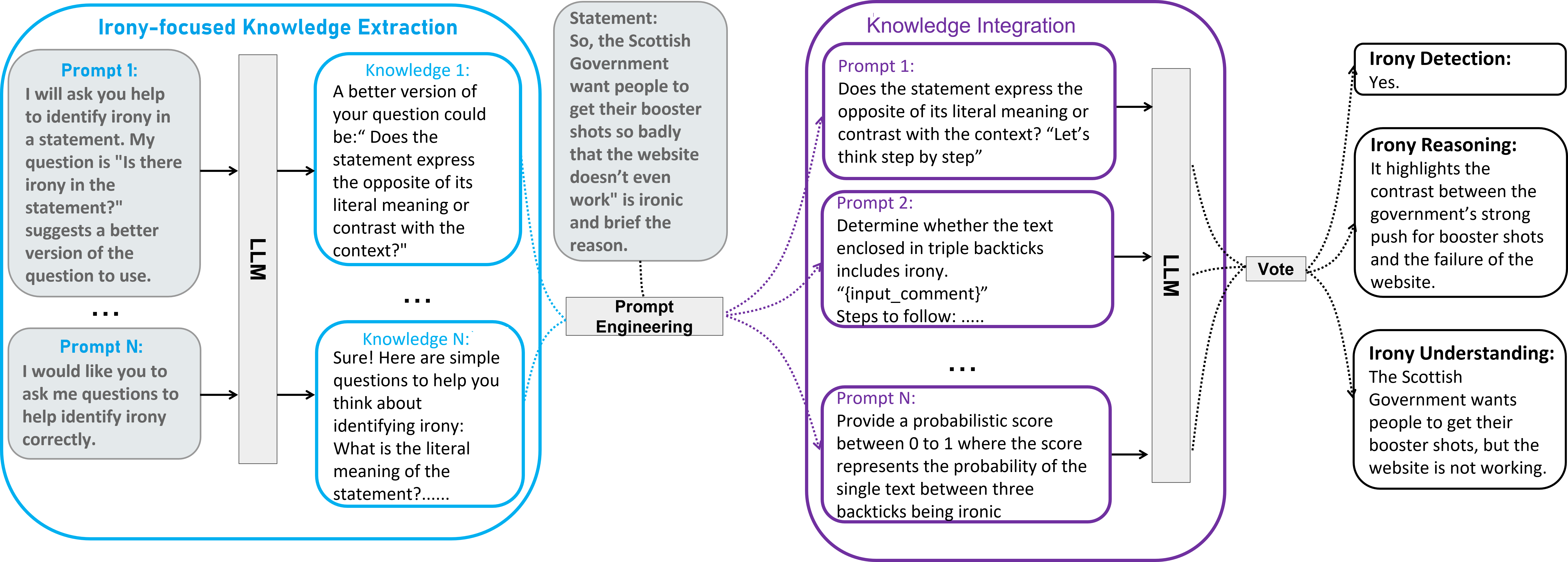}
  \caption {A concrete example of the IDADP framework for irony detection, understanding, and reasoning. In the Irony-focused Knowledge Extraction stage, multiple prompts are used to elicit different aspects of irony-related knowledge from the language model. The Knowledge Integration stage incorporates this knowledge into targeted prompts designed for specific tasks. The outputs are then aggregated using a voting mechanism to produce the final results for irony detection (binary classification), irony reasoning (explanation of contrast or context), and irony understanding (interpretation of the ironic statement).}
  \label{fig:Samples}
\end{figure*}
\IEEEPARstart{I}{rony} is a rhetorical device or figure of speech \cite{booth1974rhetoric,lear2011case} that involves expressing meaning through language that typically conveys the opposite, often for humorous or emphatic effect. 
Irony on social media often takes the form of verbal irony and situational irony. Verbal Irony occurs when someone says something but means the opposite\textcolor{blue}{\cite{joshi2017automatic}.} Situational Irony occurs when the actual outcome of a situation differs from what was expected \cite{lucariello1994situational}. Irony often relies on the broader context, 
such as
posters' personalities, cultural references, or the situation they are commenting on. The sneaky and subtle nature of irony poses a significant challenge for 
other NLP tasks, including sentiment analysis, misinformation detection, and machine translation. All these tasks need models to understand the true intent behind the text. 

Irony detection involves developing algorithms that identify and interpret irony in the text by recognizing linguistic cues like word choice, sentence structure, and context \cite{reyes2012humor}. Effective irony detection not only classifies ironic statements accurately but also explains these classifications, enhancing system transparency \cite{potamias2020transformer}. Additionally, it should also be adaptable across platforms and content types.

Irony detection is an active area of research in Natural Language Processing. 
However, existing research exhibits three limitations: \textbf{1) Generalization:} Irony detection models trained on limited or domain-specific datasets struggle to generalize across diverse real-world scenarios, including varying platforms, contexts, and cultures, thus hindering the development of robust, generalizable systems \cite{jang2024generalizable}. 
\textbf{2) Reasoning:} Accurate irony detection necessitates common-sense reasoning and real-world knowledge, 
including understanding undesirable situations and typical exaggerations. Providing transparent explanations for the model's decision-making processes remains underexplored \cite{van2018we}.
    \textbf{3) Understanding:} Current models mostly treat irony detection as a binary classification problem. 
There is a scarcity of research focused on capturing and generating sentences that accurately reflect semantic and affective intended meaning.
 This ability is crucial for effectively addressing other NLP tasks that rely on recognizing and interpreting irony \cite{farha2022semeval,van2018semeval}. 

To overcome the above limitations, we
propose a conceptual framework IDADP for irony detection, reasoning, and understanding by harnessing LLMs'
zero-shot capabilities.  LLMs has demonstrated a notable advancement in its ability to understand contextual cues and detect emotional nuances \cite{yang2024harnessing,brown2020language}, which are critical for identifying irony. The accurate interpretation of irony often hinges on the model’s capacity to discern both the literal and implicit meanings of a statement, alongside its comprehension of the broader context, including preceding dialogue elements. Concurrently, the implementation of Chain-of-Thought (CoT) prompting \cite{cheng2024chainlm,wei2022chain} on GPT has significantly enhanced its reasoning capabilities, leading to more logical and consistent approaches to problem-solving. Moreover, \textcolor{blue}{LLMs}’ sophisticated language comprehension enables it to accurately interpret complex expressions and dynamically adapt to diverse conversational styles \cite{hariri2023unlocking,ouyang2022training}. Consequently, LLMs’ integrated focus on reasoning, contextual understanding, and conversational fluency \cite{ray2023chatgpt} positions it as the optimal model for our study. \footnote{The code  is publicly available at \url{https://github.com/yipeiling/Irony}}

A concrete example (Figure \ref{fig:Samples}) is provided to illustrate the main idea of IDADP. The process begins with LLM generating irony-focused knowledge through a series of interaction prompts, as guided in \cite{white2023prompt}. This knowledge is then leveraged, in conjunction with prompt engineering techniques, to construct a variety of prompts. These enriched prompts shift \textcolor{blue}{LLM}'s attention to focus on specific aspects of irony, resulting in diverse outputs.  A voting mechanism subsequently evaluates these outputs to determine the presence of irony.  In parallel, the model provides its reasoning and generates a non-ironic sentence that retains the original statement's meaning.

 Without using any ironic training samples, this study successfully detects irony across diverse datasets and platforms while providing human-like reasoning for its conclusions.
Besides, IDADP seeks to
mitigate some of LLMs' known shortcomings, including
sensitivity to prompt engineering, inconsistent output, and the
lack of task-focused understanding \cite{azaria2024chatgpt}. The key contributions of this work can be summarized as follows: 


\begin{itemize}
\item  Firstly, we tackle the challenge of detecting irony across diverse social media platforms and formats without relying on pre-existing training datasets.
\item 
Secondly, we propose the IDADP framework to enhance LLMs' zero-shot irony detection by combining effective prompt engineering with irony-focused knowledge. Tested across six datasets, this innovative framework significantly outperforms six state-of-the-art zero-shot approaches, improving the model’s ability to detect, reason about, and understand irony.

\item Thirdly, we refine an evaluation framework that combines linguistic and contextual analysis to more accurately capture the nuances of irony.

\item 
Finally, we explore the reasoning and comprehension processes underlying LLMs’ irony prediction, highlighting existing challenges and several promising directions. This opens avenues for future research to enhance zero-shot capabilities in irony detection, reasoning, and understanding.
\end{itemize}
\section{Related work}

\subsection{Irony detection }
Irony detection requires sophisticated methods to capture the subtlety of figurative language, including recognizing sarcasm, which is often regarded as a subcategory of irony. Related works can be classified into five categories.

\subsubsection{Rule-based} This approach relies on predefined rules and patterns to identify ironic statements. For example, the authors in \cite{bharti2015parsing} provide two approaches to detect irony in Twitter's text data. The first is a lexicon generation algorithm to determine the polarity sentiment. The second detects irony based on interjection words.   Similarly, in \cite{riloff2013sarcasm}, the authors focus on identifying sarcastic tweets containing positive sentiment followed by an undesirable state. While effective for certain irony types, these methods struggle with scalability and diverse text - addressing this is the focus of this study.

\subsubsection{Lexicon-based} This method uses lexical cues and semantic patterns linked to ironic language. For instance, the authors in \cite{mladenovic2017using,gonzalez2011identifying} compare sarcastic Twitter utterances to non-sarcastic positive or negative ones. In \cite{maynard2014cares}, the researchers explore the impact of sentiment and irony in hashtags and develop a hashtag tokeniser. A key advantage is the explainability of decisions, crucial for certain tasks. However, multiple meanings of words and scalability issues can limit this approach in large-scale applications. This study aims to maintain transparency and explainability while improving scalability.

\subsubsection{Feature-based approaches}
Machine learning models for irony detection rely on linguistic features and classifiers, with their effectiveness depending on feature selection. For instance, N-grams \cite{dimovska2018sarcasm} identify recurring patterns or words indicative of irony, while part-of-speech tags \cite{fersini2015detecting} capture unusual sentence constructions. Sentiment analysis and semantic roles \cite{hernandez2015applying} help detect contradictions in meaning, and conversation history \cite{karoui2015towards} offers clues based on broader interaction context. The work \cite{kumar2023empirical} integrates these features effectively, allowing traditional models to capture intricate word-context relationships. However, these models often require complex architectures and expertise in feature selection \cite{joshi2017automatic}. While foundational and traditional methods are now often supplemented or replaced by deep learning techniques for more automatic text representation learning.

\subsubsection{Deep learning-based approaches} Unlike traditional machine learning, deep learning models learn hierarchical representations from raw text, capturing complex patterns without manual feature engineering. These models use dense vector embeddings (e.g. Word2Vec, GloVe, BERT) to capture language nuances. Neural architectures like RNNs \cite{huang2017irony}, CNNs \cite{ahuja2022transformer}, and Bidirectional LSTMs \cite{kumar2023empirical}. The work \cite{olaniyan2023utilizing} integrates these embeddings, with attention mechanisms focusing on relevant input for better irony detection in longer texts. Recent work explores transformer-based or hybrid models \cite{potamias2020transformer,belal2023leveraging}. However, complex models like transformers can be less interpretable, complicating explanations of their decision-making.

\subsubsection{Large Language Modeling approach}
Studies on using Large Language Models (LLMs) like GPT for irony detection have begun to emerge, often under zero-shot or few-shot paradigms with prompt engineering, yielding mixed results. Research \cite{gole2023sarcasm,aytekin2023generative} suggests that while GPT shows promise in other fields, it is not yet the best tool for irony detection compared to specialized models. Notably, to the best of our knowledge, no studies have specifically focused on zero-shot irony detection, inference, and understanding.

\subsection{Prompt engineering}

Prompt engineering is the process of structuring an instruction that can be interpreted and understood by a generative AI model. 
The quality of the outputs generated by an LLM is directly related to the quality of the prompts \cite{white2023prompt}. The basis of prompt engineering is rooted in developing LLMs like the GPT series \cite{radford2018improving,radford2019language,brown2020language}. Research on the GPT series has been carried out to examine how various prompt structures affect model outputs,
including the impact of length, specificity, and formatting on the quality of responses. Recent studies have focused specifically on the design optimization of prompts.

\subsubsection{Chain-of-Thought (CoT) Prompting} 
CoT introduced by \cite{wei2022chain}, improves reasoning by breaking tasks into smaller steps, similar to how humans think step by step. It can be combined with few-shot prompting for better performance on complex tasks. Building on this, the work in \cite{kojima2022large} demonstrated that LLMs can act as zero-shot reasoners by adding \textit{"Let's think step by step"} to prompts. Self-consistency \cite{wang2022self} enhances reliability by generating multiple reasoning paths, and selecting the most consistent answer. To reduce manual effort in crafting demonstrations, the work \cite{zhang2022automatic} proposed sampling diverse questions and generating reasoning chains. These methods inspired the design of our framework.

\subsubsection{Tree of Thoughts (ToT) Prompting}
ToT generalizes CoT prompting by encouraging the exploration of intermediate steps for problem-solving with language models. ToT requires defining the number of candidates and thoughts/steps for each task \cite{yao2024tree}. Similarly, authors \cite{long2023large} uses reinforcement learning instead of beam search for decision-making. The work in \cite{hulbert2023using} simplifies the approach by getting LLMs to evaluate intermediate thoughts in a single prompt. Though mainly used for arithmetic problems, we adopt the core idea: promote the solutions of the right parts, use common sense to eliminate implausible solutions,
and keep “maybe” solutions.

\subsubsection{Generated Knowledge Prompting} 
This approach involves generating knowledge from a language model and using it as additional input when answering questions. For example, the work \cite{liu2021generated} provides a representative example, consisting of two stages: First, generating question-related knowledge statements by prompting a language model. Second, integrating this knowledge to make predictions and selecting the most confident response. Our framework incorporates a similar approach.

\subsubsection{Automatic prompt generation}  
This approach focuses on \textcolor{blue}automated prompt generation to guide Large Language Models (LLMs) like GPT, replacing time-intensive manual crafting. 
Optimized Prompting (OPRO) \cite{yang2023large} uses LLMs as optimizers, generating new solutions iteratively. Similarly, Promptbreeder \cite{fernando2023promptbreeder} evolves task-prompts via an evolutionary algorithm, while APE \cite{zhou2022large} treats prompts as "programs" optimized by searching instruction candidates. Research in \cite{kong2024prewrite} trains models to rewrite under-optimized prompts. However, our experiments show that fully automated prompts are less effective for irony-focused tasks.

\subsection{Zero-shot learning in LLMs}
LLMs can generalize knowledge across domains and handle tasks without specific training or fine-tuning by using zero-shot prompts. While this is impressive, zero-shot performance is often less accurate or reliable than few-shot models \cite{brown2020language}.
 One potential reason is that, without few-shot exemplars, it is harder for models to perform well on prompts that are not similar to the format of the pre-training data. 
 To address the issue, the work in 
\cite{kojima2022large} demonstrates that LLMs are decent zero-shot reasoners by simply adding \textit{“Let’s think step by step”}. Authors in \cite{wang2023plan} replace \textit{“Let’s think step by step”} with \textit{“Let’s first understand the problem and devise a plan to solve the problem. Then, let’s carry out the plan and solve the problem step by step”} to further improve model zero-shot learning capability. FLAN (Fine-tuned Language Net) \cite{wei2021finetuned} shows fine-tuning LLMs on a collection of datasets guided by instructions significantly enhances zero-shot performance on unseen tasks. The research in \cite {sanh2021multitask} proposes a similar approach to finetune T5-11B to respond to prompts, and they also report improved performance on zero-shot learning. However, creating instruction-tuned datasets is resource-intensive. This study enhances performance through prompt optimization rather than fine-tuning with off-task data.

\section{Datasets and Challenges}
\subsection{Datasets selection}
When selecting irony detection datasets, we focus on several criteria to ensure suitability for training and evaluation, helping our models generalize across various irony types and contexts. \textbf{1) Diversity of Irony Types:} Capturing multiple irony types allows models to generalize across contexts. \textbf{2) Source Diversity:} Drawing from varied sources, platforms, or domains enhances model adaptability. \textbf{3) Annotation Quality:} Since irony is context-dependent and subtle, datasets should document annotation methods to ensure labels reflect true content. \textbf{4) Public Availability and Documentation:} Publicly available, well-documented datasets ensure reproducibility and consistency in research. Based on these criteria, six irony detection corpora were selected (Table \ref{tab:datasets_1}).


\begin{table}[htb]

  \scalebox{0.8}{
  \begin{tabular}{lcccccc}
    \toprule
    &iSarcasm\cite{farha2022semeval}&SemEval\cite{van2018semeval}&Gen\cite{oraby2017creating}&RQ\cite{oraby2017creating}&HYP\cite{oraby2017creating}&Reddit\cite{wallace2014humans}\\
    
    \midrule
    Size&1,600&4,792&6,520&1,702&1,164 &1,949 
    \\
     Length&16.4&13.7&43.3&54.2&65.3&41.35
\\
Ration&0.14&0.5&0.5&0.5&0.5&0.27 \\
  Year&2022&2018&2017&2017&2017&2014 \\

    
  Platforms&Twitter&Twitter&IAC 2.0 &IAC 2.0&IAC 2.0&Reddit\\
    Annotation&self-reported & Annotators&MTurk&MTurk&MTurk&Annotators\\

    \bottomrule
  \end{tabular}
   }
   
   \caption{
"Length" stands for the average text length within the dataset; "Ration" indicates the proportion of ironic instances relative to the total size of the dataset.}
\label{tab:datasets_1}
\end{table}


\subsection{Datasets description}

\textbf{iSarcasm \cite{farha2022semeval}} focuses on detecting intended sarcasm by minimizing labelling noise and capturing authors’ true meanings, which may differ from readers' interpretations. Contributors provide links to sarcastic and non-sarcastic tweets, implicitly labelling their texts. Trained annotators then categorize each English text by irony type, ensuring labels accurately reflect authors’ intentions.

\textbf{SemEval-2018 \cite{van2018semeval}} explores irony's effect on sentiment classification, building on challenges identified in SemEval-2014. All tweets gathered via irony-related hashtags, were manually labeled by linguistics students, with each tweet receiving a binary irony label and inter-annotator agreement checks to ensure consistency.  

\textbf{Gen \& RQ \& HYP \cite{oraby2017creating}} stem from the same research and focus on different irony types. The "Gen" dataset captures broad sarcasm instances across contexts, "RQ" highlights rhetorical questions that convey sarcasm without expecting answers, and "HYP" focuses on exaggerated statements used ironically. Each dataset, sourced from the Internet Argument Corpus (IAC 2.0), supports detailed irony analysis across diverse rhetorical forms.  

\textbf{Reddit \cite{wallace2014humans}} was collected from the online platform Reddit, where three undergraduates independently annotated each sentence with binary irony labels. This dataset reveals that context is essential for accurate irony assessment, as a sentence’s meaning can vary between ironic and non-ironic based on contextual cues.

\subsection{Challenges}

To effectively investigate the challenges inherent in this study, it is essential to first evaluate the diversity and representativeness of these datasets. Prior research \cite{jang2024generalizable} underscores the importance of cross-dataset comparisons in assessing the generalizability of irony detection models that have been fine-tuned on specific datasets. These investigations have revealed that a significant number of models struggle to generalize their performance across different datasets, suggesting that no single dataset can comprehensively capture the diverse expressions of sarcasm found in varying styles, contexts, and domains. This limitation raises critical questions regarding the efficacy of irony detection models trained on narrowly defined datasets. 

Consequently, we aim to conduct analogous preliminary experiments to further explore this issue. An experiment is set up by using 80\% of the data for fine-tuning and 20\% for testing, applying the default settings on three widely used pre-trained models for irony detection: BERT \cite{devlin2018bert}, RoBERTa \cite{liu2019roberta}, and MPNet \cite{song2020mpnet}. 

Table \ref{tab:data-diversity} illustrates that models trained on out-of-domain datasets face significant challenges. Notably, the tests conducted on the generic sarcasm (Gen), rhetorical question (RQ) and hyperbole (HYP) datasets yielded better results. However, it is essential to acknowledge that, despite these 
three
datasets containing different types of irony and varying logistical patterns, they were all sourced from the same platform and annotated by the same group of annotators. This observation suggests that, in addition to the inherent differences in irony types, the distinct nature of social media interactions and the varying methodologies of annotation may serve as 
significant
barriers to model generalization.

While these models have at least been trained on the same task, our approach intends to examine their performance in a zero-shot context, where the models are not trained on any samples 
or tasks
related to irony detection. This presents an additional layer of complexity and challenge, as we seek to understand how well these models can perform in the absence of prior exposure to irony-laden examples.

\begin{table}[h]
  \scalebox{0.8}
  {
  
  \begin{tabular}{cc|cccccc}
    \toprule
 
    Model&Cross-dataset & iSarcasm &SemEval&Gen&RQ&HYP&Reddit                \\
        \midrule
   BERT&iSarcasm  &  \textbf{0.69}&0.51&0.48&0.52&0.65&0.52                      \\
       
   &SemEval  &0.45& \textbf{0.78} & 0.50&0.61 &0.54&0.48                \\  
  
    &Gen& 0.30&0.39&\textbf{0.80}&0.78&0.64&0.53    \\    

    &RQ&0.46&0.50&0.77&\textbf{0.74}&0.76&0.57\\
   
    &HYP&0.46&0.46&0.79&0.73&\textbf{0.75}&0.55\\

    &Reddit&0.54&0.50&0.57&0.59&0.58&\textbf{0.60}\\
    \midrule
    RoBERTa&iSarcasm&  \textbf{0.78}&0.43&0.58&0.44&0.53&0.48                      \\
       
   &SemEval  & 0.55 & \textbf{0.70}&0.57 &0.59&0.56&0.55              \\  
  
    &Gen& 0.35&0.43&\textbf{0.80}&0.79&0.70&0.51    \\    

    &RQ&0.46&0.46&0.76&\textbf{0.74}&0.77&0.55\\
   
    &HYP&0.60&0.51&0.79&0.73&\textbf{0.76}&0.59\\

    &Reddit&0.51&0.58&0.60&0.53&0.58&\textbf{0.63}\\
    \midrule
    
    MPNet&iSarcasm &\textbf{0.75} &0.49&0.51&0.57&0.56&0.51                     \\
       
   &SemEval  & 0.51 &\textbf{0.74} &0.56 &0.60&0.63&0.54                \\  
  
    &Gen& 0.34&0.36&\textbf{0.80}&0.67&0.76&0.53    \\    

    &RQ&0.48&0.47&0.76&\textbf{0.74}&0.76&0.58\\
   
    &HYP&0.63&0.47&0.76&0.69&\textbf{0.70}&0.56\\

    &Reddit&0.50&0.53&0.55&0.53&0.41&\textbf{0.65}\\
    \bottomrule
    
  \end{tabular}
  }
  \caption{Performance of BERT, RoBERTa, and MPNet models on Cross-dataset irony classification.}
  \label{tab:data-diversity}
\end{table}

\section{Methodology}
In this section, we start by defining the research problem, followed by an overview of the overall framework of our model. Next, we provide a detailed explanation of the proposed method, concluding with an outline of the application process.

\subsection{Problem definition}
The key notations used throughout the article, as outlined in Table \ref{tab:notation}, followed by a formalization of the research problems from multiple perspectives.
\begin{table}[htb]

  \begin{tabular}{lll}
    \toprule
    Notation&&Description \\
    \midrule
    
    T&&Text input(written or spoken)\\
    D&& Irony-focused information
    \\
     P&& Prompt
    \\
    C&&Binary Classification function
    \\
   R&&Description of the Reasoning Process
   \\
   G&& Generalization function \\
   F&& Reasoning function \\
   U&& Understanding function 
    \\
    $D^{test}$&&Testing data\\
     E&&The expected value on the unseen test data. \\
     $M^i$&&Intended meaning. \\
     $M^l$&&Literal meaning. \\
    $f^{\theta}$&& The model with parameters $\theta$
    \\
    I&&Ironic
    \\
    N&&Non-Ironic
    \\
    \bottomrule
  \end{tabular}
  \caption{Notations and Descriptions for IDADP.}
  \label{tab:notation}
  
\end{table}

\textbf{Irony detection:} Irony detection, in computational terms, is the process of the model automatically identifying and understanding instances of irony in written or spoken language \cite{zeng2022survey}. Irony detection in the study can be formulated as a binary classification problem: 

\[
C(T) \in \{I, N\}
\]
The goal of the model is to accurately classify $T$ based on its understanding of irony.



\textbf{Generalization:} The generalization of a model refers to its ability to perform well on new, unseen data that was not included in the training set\cite{shalev2014understanding}. The generalization of a model can be expressed as:

\[
{G(T)} = D^{\text{test}}[f^{\theta}(T)] - E(T)
\]

Instead of relying on specific examples for each task, the model draws on its understanding of language, context, and general knowledge to infer appropriate responses from the input.


\textbf{Reasoning:} Reasoning refers to the model's ability to process information, draw inferences, and provide conclusions or answers based on the given input \cite{duan2020machine}.

\[
R(T) = F(P(I, T))
\]

The reasoning process is successful if the model generates a logically sound conclusion and a human-readable and understanding reason.

\textbf{Understanding:} Understanding in irony detection 
 refers to the model's ability to recognize and interpret statements whose intended meaning is different or opposite from the literal meaning \cite{bruntsch2017studying}. 

\[
M^{i} \neq M^{l}
\]


The study indicates that the model generates sentences that do not contain irony but retain the same intended meaning, which can be described as:

\[
U(T) = M^{i} \quad \text{and} \quad U(T) \not\in I^{n}
\]
\subsection{Theory analysis}
 In-Context Learning \cite{brown2020language} relies on well-constructed prompts to "teach" the model how to handle the task by providing it with well-constructed prompts or examples during the process. 
However, the performance of in-context learning is: \textbf{1) Highly sensitive} to the quality and format of the prompt. Even minor adjustments in wording or the order of examples can result in significant performance variations. For instance, a prompt that is structured to highlight the contrast between literal and intended meanings can enhance the model's ability to detect irony effectively. Conversely, vague or poorly framed prompts may confuse the model, leading to misclassification.
\textbf{2) Less reliable} for complex tasks such as irony detection, which require deep task-specific knowledge. This is particularly true for understanding various pattern of irony that the model may not have encountered during its pre-training phase. \textbf{3) Inherently variable }is the nature of GPT models, which involves predicting the next word based on probabilities. This randomness can result in fluctuations in the generated text, which presents additional challenges for irony reasoning and understanding. 


The following theoretical analysis reveals the insights of IDADP, leading to mitigating known In-Context Learning limitations through prompt optimization.

\subsubsection{Attention Mechanism in Transformers (Highly sensitive)}
Transformers, the architecture underlying large language models (LLMs), utilize an attention mechanism that enables the model to focus on different parts of the input sequence more effectively.
Prompts designed in various ways activate the model's learned patterns, guiding it toward specific types of outputs. One of the more advanced techniques is self-consistency \cite{wang2022self}, which addresses the limitations of the traditional greedy decoding approach used in Chain-of-Thought (CoT) prompting. Self-consistency involves sampling multiple, diverse reasoning paths through few-shot CoT and using these generated outputs to identify the most consistent answer. Inspired by this approach, we employ a variety of prompt designs with LLMs to enhance the efficacy and robustness of the prompting process.

\subsubsection{Biasing the Model via Contextual Cues. (Less reliable)} Language models rely on probabilistic methods to generate text. 
The structure and wording of a prompt heavily influence which tokens are deemed more probable, essentially "biasing" the model toward certain responses. By generating task-specific prompt, we can bias the model toward generating responses that align with our task. Specific irony-focused keywords and instructions can improve the likelihood of generating relevant, high-quality responses. For the zero-shot setting, we can't give the samples but we can give relevant knowledge to clear and specific prompts to guide the model to understand the task based solely on the instructions. Thus, our framework aims to generate knowledge prompts by asking model questions related to the task, allowing us to extract key terms and essential instructions.
\begin{figure*}[tbh]
  \centering
  \includegraphics[width=\linewidth]
  {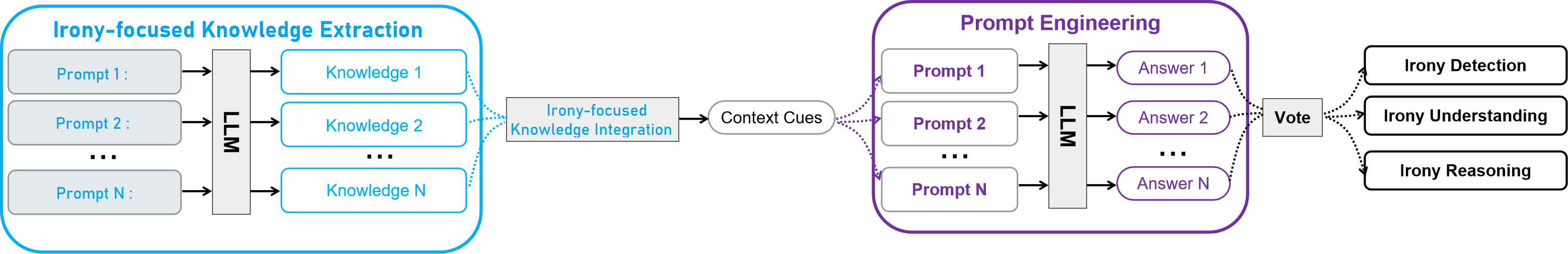}
  
  \caption {The overall architecture of the IDADP framework. The process begins with Irony-focused Knowledge Extraction, where multiple tailored prompts are used to extract diverse irony-related knowledge from the language model (LLM). This extracted knowledge is then integrated and passed as context cues into the Prompt Engineering module, where additional prompts generate candidate answers for irony detection, understanding, and reasoning tasks. The final outputs are determined through a voting mechanism that aggregates the answers from multiple prompts, providing robust predictions and generation for each downstream task.}
  \label{fig:architecture}
\end{figure*}

\subsubsection{Zero-shot and Contextual Framing (Inherently variable)}Large Language models are pre-trained on a broad corpus of data using unsupervised learning, where they learn general language patterns. However, many models struggle with task-specific knowledge in zero-shot learning tasks. Without examples or fine-tuning, it may generate incorrect or overly general responses. By framing the prompt with more explicit context and instructions, we can influence the model’s "thinking" (its probabilistic output) and guide it toward a desired style or form of response. 

\subsection{Overall Framework}
The overarching structure of the proposed IDADP model is depicted in Figure \ref{fig:architecture}.

\subsubsection {Irony-focused Knowledge Extraction}
This phase aims to prompt LLM to generate comprehensive, irony-focused knowledge rather than relying on domain experts or predefined datasets. This knowledge will be abstracted at the next phase as context cues to help machines do zero-shot learning. These prompts all are generated by certain patterns guided by \cite{white2023prompt}. These patterns describe effective techniques for accomplishing different interaction objectives.   

\textbf{The Flipped Interaction Pattern:} The pattern guides the model in generating more accurate questions to initiate the conversation. For example, \textit{“I would like you to ask me questions to identify irony correctly."}
    
\textbf{The Persona Pattern:} This pattern allows the model to assist us when the exact details of the required inputs are unknown. For example, \textit{“Act as an annotator to label irony datasets."}
    
\textbf{The Question Refinement Pattern:} When we lack domain expertise, it can be challenging to phrase a question effectively or include all the relevant details. This pattern enables the model to generate more professional questions, leading to more accurate responses. For example, \textit{“I will ask your help to identify irony in a statement. My question is `Is there irony in the statement?' suggests a better version of the question to use."}
    
\textbf{The Recipe Pattern:} The Recipe pattern prompts the model to produce structured, clear prompts. For example, \ \textit {provide a complete sequence of steps to identify an irony in a statement.}

\subsubsection {Irony-focused Knowledge Integration} This phase integrates irony-focused knowledge from phase 1 with general contextual cues. Precise and detailed prompts are crucial for optimal model performance, ensuring relevance to the task. We incorporate three knowledge types: the definition of irony, its nuanced nature, and the process of irony detection.

\textbf{Simple but specific definition of irony with one sentence:} The first type aims to elicit quick, intuitive, or pattern-recognition-based answers from the model, focusing on straightforward and specific definitions of irony. For example, \textit{irony expresses the opposite of its literal meaning or contrast with the context.}
    
\textbf{Irony detection specificity features:} The type prompt emphasizes the importance of features in the understanding of ironic statements. For example, \textbf{discrepancy:} between what is said and what is meant. and \textbf{ contrast:} between expectation and reality presented in the statement.

\textbf{The process of irony detection:} This type of knowledge is specifically related to the irony detection process. For example, begin by asking model to assess whether a statement is ironic, using a prompt such as \textit{“Is the following statement ironic?"}. Next, provide the statement along with relevant context to enhance understanding. For example, \textit{“A person says, `I love waiting in line for hours!' after spending three hours at the DMV."} Then, prompt the model to identify the literal meaning of the statement: \textit{“What is the literal meaning?"} Following this, encourage the model to evaluate any discrepancies between the literal meaning and the situation by asking: \textit{“Does the literal meaning match the actual situation?"} Finally, conclude by asking the model to determine whether the statement is ironic based on the previous analyses.

\subsubsection {Prompt Engineering}

This phase organizes all provided content, employing advanced prompt engineering techniques to generate varied prompts, including:

\textbf{
Task-Specific Chain-of-Thought (CoT)}  unlike standard zero-shot CoT, which uses instruction \textit{“step-by-step"} to guide the model in generating its reasoning process, we incorporate task-specific guidelines to prompt more precise reasoning. For example, the model's reasoning roadmap is controlled by specifying a sequence of required steps. These steps are generated through the application of the meta prompting. 

\textbf{Meta Prompting} is previously demonstrated in the domain of mathematical problem-solving \cite{zhang2023meta}. We adopt the method to guide a more abstract and structured interaction with LLMs that focuses on form and pattern rather than traditional content-centric methods.

\textbf{Probabilistic Classification} not only predicts class labels but also quantifies uncertainty, providing valuable insights for decision-making in critical applications. It is especially effective for sarcasm detection due to the subtle and nuanced nature of irony.
    
\subsubsection {Vote}
Voting is a rule-based classification method that determines outcomes based on results from multiple prompts, with rules varying depending on the target criteria. In our case, we apply a majority-consent rule (e.g., best of three) to reach a decision.

\section {Experiments}
This section conducts a quantitative evaluation to assess the effectiveness of IDADP and provides a comprehensive understanding of the performance of the model.

\subsection{Settings}

In this study, we employed both the GPT-3.5-turbo and Gemini-1.5-flash. GPT-3.5-turbo is a variant of the GPT-3.5 family. Since state-of-the-art prompting techniques have been heavily influenced by GPT-3, conducting a comparative evaluation on the same base model helps distinguish whether the enhancements of our method stem from technical innovations or the inherent capabilities of the base model. Additionally, GPT-3.5 is more accessible and cost-effective compared to GPT-4, making it a practical choice for large-scale experiments and reproducibility. To minimize variability, we adhered to the default parameter settings from OpenAI’s API, setting only \textit{max\_tokens} to 300 and \textit{temperature} to 0.3.

Similarly, Gemini-1.5-flash, developed by Google, represents one of the latest advancements in large language models and is designed to provide high-speed, cost-efficient inference without significant sacrifices in model performance. Its architecture and deployment through Google’s AI platform make it an appealing alternative for benchmarking alongside GPT-based models. For consistency and fair comparison, we used Gemini-1.5-flash with its default API parameter settings, matching the maximum output length (\textit{max\_tokens}: 300) and temperature (0.3) to those used with GPT-3.5-turbo. This approach ensures that our evaluation focuses on methodological improvements rather than discrepancies in model configuration or resource allocation.

Handling models output formatting issues involves several steps in the experiment: \textit{1)} Clarify and standardize the output in JSON format within the prompt.
\textit{2)} Abstract numbers or special information using string regularization techniques.
\textit{3)} Post-process the output to standardize and filter out empty output.

\subsection{Baselines}
\subsubsection{Fine-Tuned Models}
To better contextualize the gap
between zero-shot and supervised learning approaches, we compare IDADP with three fine-tuned irony detection models: BERT \cite{devlin2018bert}, RoBERTa \cite{liu2019roberta} and MPNet \cite{song2020mpnet} under zero-shooting and fine-tuning settings. 

\subsubsection{GPT3.5 \& Gemini1.5}
The two baseline is involved to help assess whether performance improvements stem from IDADP itself or the underlying model’s capabilities. During inquiring output, just simply ask GPT3.5 and Gemini1.5 \textit{“Determine whether [input comment] includes irony."}

\subsubsection{Zero-shot CoT \cite{kojima2022large}}Zero-shot CoT Prompting enables complex reasoning capabilities through intermediate reasoning steps by adding \textit{“Let's think step by step"} to the original prompt.
As demonstrated in Table \ref{tab:prompt_cot} \begin{table}[h]
 \centering
 \scalebox{0.9}{
 \centering
  
   \begin{tabular}{l}
     \toprule
    Determine whether $[input\_comment]$ include irony.                 \\
       
 Steps to follow:\\
 1. Let's think step by step.\\
 2. Please write the reason why you think this statement has irony.\\
 3. Please rephrase this statement without the irony with a new line.\\
 4. the result in only a JSON format where the key is "irony"\\ and the value is  1 for irony, 0 for No-irony.\\
     \bottomrule
   \end{tabular}
 }
 \caption{Prompt Sample of Zero-Shot CoT}
  \label{tab:prompt_cot}
 \end{table}

\subsubsection{Auto-CoT \cite{zhang2022automatic}}
\begin{table}[h]
 \centering
 \scalebox{0.9}{
 \centering
    \begin{tabular}{l}
     \toprule
    Determine whether $[input\_comment]$ include irony.                 \\
       
 Steps to follow:\\
 1. Study the following samples.\\
 2. [Example 1]\\
 3. [Example 2]\\
 4. [Example 3]\\
 5. [Example 4]\\
 6. [Example 5]\\
 7. [Example 6]\\
 8. Let’s think step by step.\\
 9. Please write the reason why you think this statement has irony.\\
 10. Please rephrase this statement without the irony with a new line.\\
 12. the result in only a JSON format where the key is "irony"\\ and the value is 1 for irony, 0 for No-irony.\\
     \bottomrule
   \end{tabular}
 }
 \caption{Prompt sample of Auto-CoT}
  \label{tab:prompt_autocot}
 \end{table}
Consists of two stages: \textbf{1) question clustering}, which groups dataset questions by similarity to streamline varied question types, and \textbf{2) demonstration sampling}, where a representative question from each cluster is selected, with its reasoning chain generated using Zero-Shot CoT. Simple heuristics ensure coherence in the reasoning chain. Unlike traditional clustering, we select samples directly from each dataset, enhancing flexibility and better reflecting dataset diversity and complexity. As demonstrated in Table \ref{tab:prompt_autocot}. 

\subsubsection{APE \cite{zhou2022large}}
Automatic prompt engineer (APE) is a framework for automatic instruction generation and selection.  they discover a general prompt \textit{“Let’s work this out in a step-by-step way to be sure we have the right answer."} can improve performance on different datasets and tasks. As demonstrated in Table \ref{tab:prompt_ape}.

 \begin{table}[h]
 \centering
 \scalebox{0.9}{
 \centering
  
   \begin{tabular}{l}
     \toprule
    Determine whether $[input\_comment]$ include irony.                 \\
       
 Steps to follow:\\
 1. Let’s work this out in a step-by-step way\\ to be sure we have the right answer.\\
 2. Please write the reason why you think this statement has irony.\\
 3. Please rephrase this statement without the irony with a new line.\\
 4. the result in only a JSON format where the key is "irony"\\ and the value is  1 for irony, 0 for No-irony.\\
     \bottomrule
   \end{tabular}
 }
 \caption{Prompt sample of APE}
  \label{tab:prompt_ape}
 \end{table}

\subsubsection{PS \cite{wang2024plan}} Plan-and-Solve (PS) Prompting expand In Zero-shot CoT includes the instructions \textit{“Let’s first understand the problem and
devise a plan to solve the problem. Then, let’s carry out the plan and solve the problem step by step"}. As demonstrated in Table \ref{tab:prompt_ps}. 
 \begin{table}[h]
 \centering
 \scalebox{0.9}{
 \centering
  
   \begin{tabular}{l}
     \toprule
    Determine whether $[input\_comment]$ include irony.                 \\
       
 Steps to follow:\\
 1. Let’s first understand the problem and devise a plan \\to solve the problem\\
 2. let’s carry out the plan and solve the problem step by step.\\
 3. Please write the reason why you think this statement has irony.\\
 4. Please rephrase this statement without the irony with a new line.\\
 5. the result in only a JSON format where the key is "irony"\\ and the value is  1 for irony, 0 for No-irony.\\
     \bottomrule
   \end{tabular}
 }
 \caption{Prompt sample of PS}
  \label{tab:prompt_ps}
 \end{table}

\subsubsection{PS+ \cite{wang2024plan}}
PS+ aims to reduce errors caused by missing necessary reasoning steps, such as “extracting relevant variables and their corresponding numerals," explicitly instructing model not to overlook important information in the input problem statement. As demonstrated in Table \ref{tab:prompt_ps+}. 

 \begin{table}[h]
 \centering
 \scalebox{0.9}{
 \centering
  
   \begin{tabular}{l}
     \toprule
    Determine whether $[input\_comment]$ include irony.                 \\
       
 Steps to follow:\\
 1. Let’s first understand the problem and check if contains a\\ discrepancy between what is said and what is meant\\
 2. let’s carry out the plan and pay attention to finding ironic \\ words or phases.\\
 3. solve the problem step by step.\\
 3. Please write the reason why you think this statement has irony.\\
 4. Please rephrase this statement without the irony with a new line.\\
 5. the result in only a JSON format where the key is "irony"\\ and the value is  1 for irony, 0 for No-irony.\\
     \bottomrule
   \end{tabular}
 }
 \caption{Prompt sample of PS+}
  \label{tab:prompt_ps+}
 \end{table}

\subsection{Prompts in IDADP}

In this section, we will give three prompt samples (Table \ref{tab:prompts_IDADP}) generated from the IDADP framework and used in our experiments. 

\begin{table}[h]
\centering
\scalebox{0.9}{
\centering
  
  \begin{tabular}{l}
    \toprule
    
   Determine whethe $[input\_comment]$ include irony.\\ Let's think step by step               \\
 \midrule      
Sample 1: Steps to follow:\\
1. Identify the irony: Determine which part of the sentence\\ conveys the opposite of what is meant.\\
2. Clarify the intent: Express the actual meaning directly\\
3. Please write the reason why you think this statement has irony.\\
4. Please rephrase this statement without the irony with a new line.\\
5. the result in only a JSON format where the key is "irony"\\ and the value is  1 for irony, 0 for No-irony.\\

        \midrule
Sample 2: Steps to follow:\\
1. The text is not ironic if the statement does not contain\\ a discrepancy between what is said and what is meant.\\
2. The text is not ironic if There is no unexpected outcome or contrast\\ between expectation and reality presented in the statement.\\
3. Please write the reason why you think this statement has irony.\\
4. Please rephrase this statement without the irony with a new line .\\5. the result in only a JSON format where the key is "irony"\\ and the value is  1 for irony, 0 for No-irony.\\

        \midrule
Sample 3: Steps to follow:\\
1. Please provide a probabilistic score ranging from 0 to 1,\\ representing the likelihood that the text is ironic.\\
2. The threshold for irony detection is set to 0.7.\\
3. Please write the reason why you think this statement has irony.\\
4. Please rephrase this statement without the irony with a new line.\\
5. the result in only a JSON format where the key is "irony"\\ and the value is  1 for irony, 0 for No-irony.\\
    \bottomrule
  \end{tabular}
}
\caption{Three prompt samples of IDADP}
 \label{tab:prompts_IDADP}
\end{table}

\subsection{Evaluation}
This section introduces evaluation metrics and strategies for assessing IDADP's irony detection, reasoning, and comprehension.  These metrics provide unique perspectives on the model's effectiveness across diverse contexts, enabling a comprehensive performance evaluation.

\subsubsection{Detection}
We evaluated IDADP's performance in irony detection using three metrics across six datasets: \textbf{1) Precision (P), 2) Recall (R), and 3) Micro-average F1-score (F)}. Precision measures the model's accuracy in correctly identifying ironic statements without misclassifying non-ironic ones, though it doesn’t capture missed ironic cases (false negatives). Recall assesses the model’s ability to identify all ironic instances, with high recall indicating strong identification but potentially more false positives. The micro-average F1-score provides equal weight to all classes, making it suitable for imbalanced datasets like iSarcasm and Reddit, as it prevents dominant classes from overshadowing minority class performance, offering a balanced view of overall effectiveness.

\subsubsection{Reasoning}
The reasoning process is deemed successful if the model generates a logically sound conclusion along with a human-readable and understandable explanation.  
We evaluated the model’s reasoning 
using four metrics:\textbf{ 1) Flesch-Kincaid readability score (F); 2) Human evaluators Score (H); 3) Standard Deviation(S); 4) B Measure(B) }

The Flesch-Kincaid readability score \cite{farr1951simplification}, based on sentence length and word complexity, is widely used in industries like education and digital content. Higher scores indicate easier readability, while lower scores mark more difficult passages. However, the Flesch-Kincaid readability score overlooks meaning and context,  it provides limited insight into overall readability. 

Therefore, human evaluation is essential for a comprehensive assessment of the study. Table \ref{tab:human_reasoning} illustrates how human evaluators scored the reasoning outputs. Evaluator's criteria are given to the following: \textbf{1) Contextual Accuracy (1 point):} Does the response refer to the correct context?
\textbf{2) Internal Consistency (1 point):} Is the reasoning in the response free of contradictions?
\textbf{3) Clarity of Structure (1 point):}  Is it structured in a way that the main point is introduced, elaborated, and concluded logically?
To manage larger datasets within resource constraints, Three graduate students with varying levels of familiarity with natural language processing (NLP) tasks are assigned to evaluate a subset of the data, specifically one-third of the total dataset. Before starting the full evaluation, each student is assigned 10 samples, which are judged and adjusted by a professional. By dividing the work into manageable portions, we ensure an efficient distribution of workload among evaluators, preventing any individual from becoming overwhelmed.

\begin{table}[h]
\centering
\scalebox{0.9}{
\centering
    \begin{tabular}{ll}
    \toprule
   \textbf{Original statement:}&\textit{“What a beautiful day for a walk," }                \\
   \textbf{}&\textit{ said as it rains heavily.}                \\
   \midrule
     Model\_1 response:&“The speaker is using irony because the weather\\& is clearly bad, making the compliment about\\& the day’s beauty an obvious contradiction."\\
    Annotation: &\textbf{3 points}\\
    Remarks:&The response correctly explains the contradiction \\&between the literal and intended meaning.\\
    \midrule
     Model\_2 response:&“Basically, the thing is that irony happens when\\& you say something but don't really mean it, 
    \\&which is kind of what’s going on here. \\&Like, they said something good about the weather \\&but it’s raining, so it's ironic."\\
    Annotation: &\textbf{2 points}\\
    Remarks: &Somewhat clear, but could be better organized\\& or more concise.\\
     \midrule
    Model\_3 response:&“The statement seems ironic, but in reality, it \\&may not be because the weather could actually \\&be pleasant depending on someone’s preferences, \\&even if it's raining"
    \\
    Annotation: &\textbf{1 points}\\
    Remarks: &The structure is not bad, but the reasoning is\\& flawed, as it fails to properly identify irony.\\
    
    \bottomrule
  \end{tabular}
}
\caption{Examples of human annotation} 
 \label{tab:human_reasoning}
\end{table}
To assess the consistency of readability in the models' responses, we calculated the standard deviation of the Flesch-Kincaid readability scores. A low standard deviation indicates greater consistency in readability, suggesting better predictability and stability in response quality. Conversely, a high standard deviation suggests that a model's responses vary considerably in readability, with some responses being highly readable while others are significantly less so.

Standard deviation is used to assess the stability of the Flesch-Kincaid readability score.  A high standard deviation generally indicates that a prompt
may produce responses that vary greatly in terms of readability, with some responses being very readable and others much less so. Low standard deviation suggests more consistency
in readability, which is a positive sign for predictability and
stability in response quality.

The B measure balances the Flesch-Kincaid readability score with human evaluators' scores, ensuring a comprehensive evaluation of the model’s reasoning by considering readability, logic, and correctness. The formula is:
        B =
\[
\frac{\text{Flesch-Kincaid Readability Score}}{100} + \frac{\text{Human Evaluators' Score}}{3}
\]

\subsubsection{Understanding}
Evaluating irony understanding in language models is challenging, as it requires grasping context, intention, and the contrast between literal meaning and actual intent. In this study, we assess irony understanding through an Irony rephrasing task, where the model rephrases ironic sentences in a non-ironic manner while preserving the intended meaning. 
We use Cosine Similarity with BERT sentence embeddings to measure the similarity between a no-ironic sentence rephrased by model and the intended meaning provided by the author.
The resulting score ranges from 0 to 1, with 1 indicating the model's perfect understanding of the intended meaning of the ironic sentence.

\begin{table*}[htb]
 \begin{center}
 \resizebox{\textwidth}{!}{
  \begin{tabular}{l|ccc|ccc|ccc|ccc|ccc|ccc|ccc}
  \toprule
   &\multicolumn{3}{c|}{Overall(Mean)}&\multicolumn{3}{c|}{iSarcasm}&\multicolumn{3}{c|}{SemEval}&\multicolumn{3}{c|}{Gen}&\multicolumn{3}{c|}{RQ}&\multicolumn{3}{c|}{HYP}&\multicolumn{3}{c}{Reddit}\\
  \midrule
  \midrule{Detection}&P&R&F1&P&R&F1&P&R&F1&P&R&F1&P&R&F1&P&R&F1&P&R&F1\\
   \midrule
    \multicolumn{22}{c}{Fine-tuning Models (80\% training data)}\\
    \midrule
    Mpnet &0.75&0.73&0.73&0.78&0.76&0.75&0.78&0.75&0.74&0.85&0.80&0.81&0.74&0.74&0.74&0.72 &0.71&0.70&0.65&0.65&0.65\\
     Roberta &0.73&0.74&0.74&0.79 &0.78 &0.78&0.70&0.70&0.70&0.80&0.80&0.80&0.74&0.74&0.74&0.70 &0.79 &0.76&0.64&0.62&0.63\\
      Bert &0.74&0.75&0.73&0.70&0.69&0.69&0.79&0.78&0.78&0.80&0.80&0.80&0.76&0.75&0.75&0.76&0.75&0.75&0.61&0.71&0.60\\
    \midrule
    \midrule
    \multicolumn{22}{c}{Zero-shot GPT base Models (No training data)}\\
    \midrule
    GPT3.5
    &0.64&0.57&0.45&0.58&0.63&0.39&0.50&0.50&0.42&0.68&0.59&0.54&0.74&0.57&0.49&0.70&0.56&0.45&0.62&0.58&0.43\\
     Gemini1.5
    &0.59&0.55&0.48&0.54&0.60&0.36&0.54&0.50&0.43&0.44&0.48&0.34&0.58&0.53&0.50&0.79&0.56&0.47&0.66&0.63&0.45\\
    Zero-CoT &0.60&0.61&0.58&0.60&0.71&0.58&0.47&0.47&0.47&0.61&0.61&0.61&0.64&0.63&0.63&0.67&0.66&0.64&0.58&0.59&0.57\\
    Auto-CoT
    &0.62&0.52&0.31&0.57&0.54&0.20&0.49&0.50&0.35&0.58&0.51&0.36&0.77&0.51&0.31&0.74&0.52&0.36&0.59&0.52&0.28\\
     APE
    &0.64&0.65&0.63&0.60&0.69&0.58&0.62&0.60&0.59&0.65&0.64&0.64&0.69&0.70&0.69&0.68&0.68&0.68&0.59&0.60&0.58\\
     PS
    &0.57&0.56&0.50&0.62&0.59&0.60&0.54&0.51&0.40&0.62&0.53&0.42&0.62&0.62&0.55&0.47&0.56&0.52&0.57&0.53&0.52\\
    PS+
    &0.63&0.53&0.37&0.63&0.56&0.57&0.49&0.50&0.35&0.58&0.51&0.36&0.77&0.51&0.31&0.74&0.52&0.36&0.59&0.52&0.28\\
    
\midrule

\textbf{IDADP\_GPT} &\textbf{0.71}&\textbf{0.71}&\textbf{0.71}& \textbf{0.65} & \textbf{0.72} & \textbf{0.67} & \textbf{0.65} & \textbf{0.65} & \textbf{0.65} & \textbf{0.74} & \textbf{0.74} & \textbf{0.74} & \textbf{0.80} & \textbf{0.73} & \textbf{\underline{0.82}} & \textbf{0.81} & \textbf{0.78} & \textbf{0.76} & \textbf{0.63} & \textbf{0.64} & \textbf{0.64} \\

\textbf{IDADP\_Gemini} &\textbf{0.70}&\textbf{0.70}&\textbf{0.71}& \textbf{0.67} & \textbf{0.67} & \textbf{0.67} & \textbf{0.65} & \textbf{0.65} & \textbf{0.65} & \textbf{0.74} & \textbf{0.74} & \textbf{0.74} & \textbf{0.69} & \textbf{0.85} & \textbf{\underline{0.76}} & \textbf{0.85} & \textbf{0.84} & \textbf{\underline{0.85}} & \textbf{0.60} & \textbf{0.63} & \textbf{0.60} \\

   \bottomrule
 \end{tabular}
 } 
 \end{center}
 \caption{Results of the IDADP in comparison to the baselines across six datasets. P: Precision; R: Recall; F1: Micro F1; The best performances are highlighted in bold.}
 \label{tab:detection-results}
\end{table*}
\subsection{Results }
This section presents the experiments’ results to address the limitations driving our study.

\begin{table*}[h]
\centering
\centering
  
  \begin{tabular}{lllll}
    \toprule
    Index&Posts & Ironic&Prediction &Human Evaluate                 \\
        \midrule
   1&“villainous pro tip: change the device name on her &&&\\
    &Bluetooth devices so she doesn’t forget u"&No&Yes&Yes \\

\textit{Human Reason:}&\multicolumn{4}{l}{\textit{the advice is villainous, implying bad or unethical behaviour }}\\
&\multicolumn{4}{l}{\textit{rather than something genuinely helpful.}}
\\
  \midrule 
 
    2&“Where did they get 1.22xg from without the pen?!" 
   &No&Yes&No \\   
  
\textit{Human Reason:}&\multicolumn{4}{l}{\textit{It appears to be a straightforward question or expression }}\\
&\multicolumn{4}{l}{\textit{ of confusion about the expected goals (xG) statistic in sports}}\\

\midrule 
 
    3&“Sort of a `Read my lips, no new taxes.' GHW Bush moment 
   &No&Yes&Not Sure \\  
   &for Obama."
   &&& \\  
  
\textit{Human Reason:}&\multicolumn{4}{l}{\textit{the reference to Bush's famous failed pledge adds an ironic twist}}\\
&\multicolumn{4}{l}{\textit{to Obama's situation }}
\\
\midrule 
  4&“So much for the Hippocratic oath"
   &Yes&No&Yes \\   
  
\textit{Human Reason:}&\multicolumn{4}{l}{\textit{it typically implies that a doctor or medical professional has }}\\
&\multicolumn{4}{l}{\textit{acted in a way 
 that contradicts the principles of the Oath}}\\
\midrule 

 5&“Would they take it back?" 
   &Yes&No&Not Sure \\   
  \textit{Human Reason:}&\multicolumn{4}{l}{\textit{Without context, it may just be a straightforward question.}}\\
\midrule

6&“No, their primary motivation is to get money from   &Yes&No&No \\   
& the wealthy then do what the wealthy want" &&& \\
\textit{Human Reason:}&\multicolumn{4}{l}{\textit{It appears to be a direct, critical commentary on the}}\\
&\multicolumn{4}{l}{\textit{ motivations of a group or organization}}\\

    \bottomrule
  \end{tabular}
\caption{misclassification samples along with their predicted labels, human evaluations and human reason}
 \label{tab:misclass}
\end{table*}

\begin{figure*}[tbh]
  \centering
  \includegraphics[width=\linewidth,trim=8 4 8 4,clip]
  {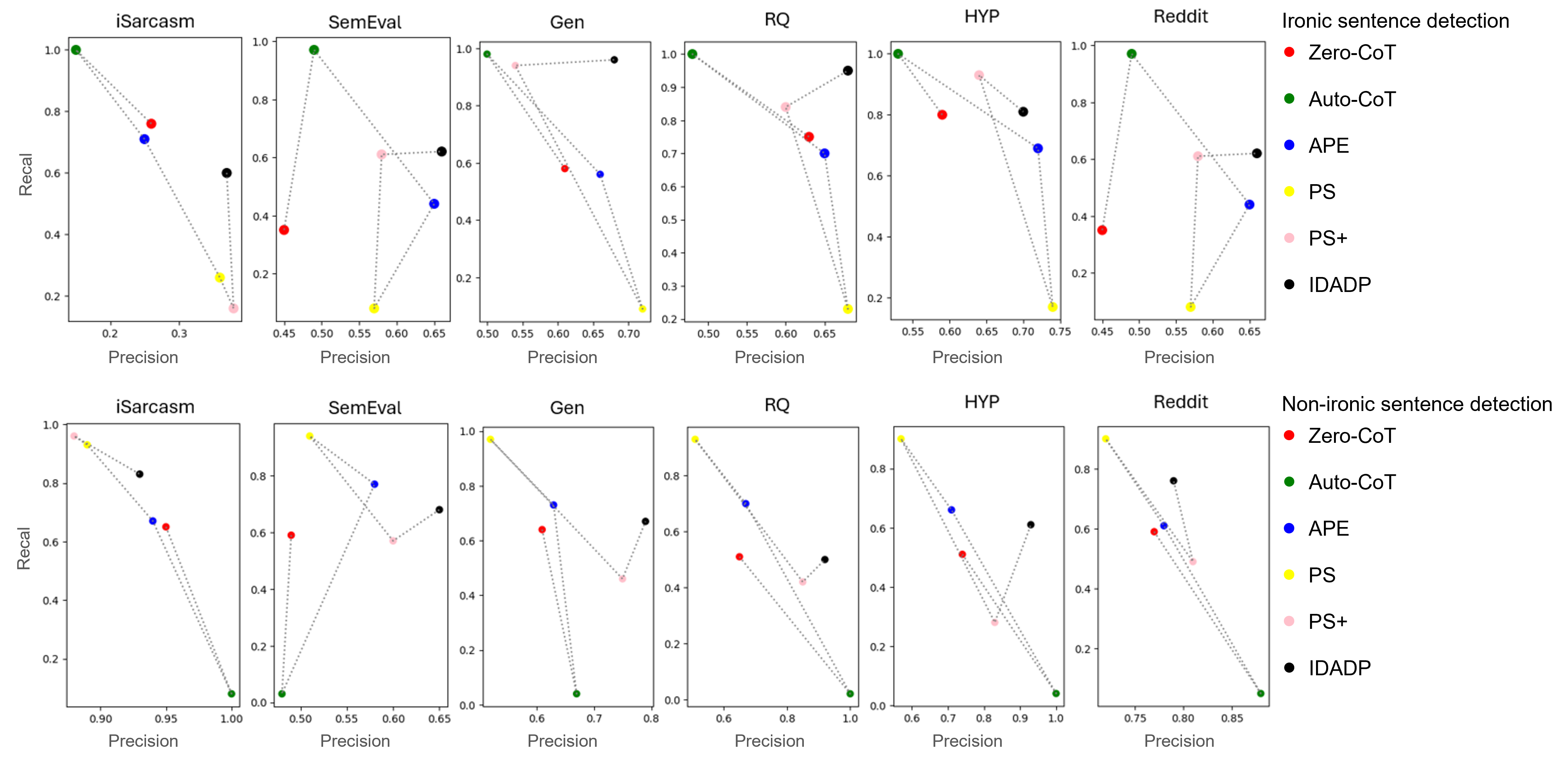}
  \caption {Performance comparison of IDADP and baseline models on ironic and non-ironic sentence 
 detection across six datasets.}
  \label{fig:positive}
\end{figure*}

\begin{table*}[htb]
 \begin{center}
 \resizebox{\textwidth}{!}{
  \begin{tabular}{l|cccc|cccc|cccc|cccc|cccc|cccc}
  \toprule
   &\multicolumn{4}{c|}{iSarcasm}&\multicolumn{4}{c|}{SemEval}&\multicolumn{4}{c|}{Gen}&\multicolumn{4}{c|}{RQ}&\multicolumn{4}{c|}{HYP}&\multicolumn{4}{c}{Reddit}\\
  \midrule
   
   {Reasoning}&F$\uparrow$&S$\downarrow$&H$\uparrow$&B$\uparrow$&F$\uparrow$&S$\downarrow$&H$\uparrow$&B$\uparrow$&F$\uparrow$&S$\downarrow$&H$\uparrow$&B$\uparrow$&F$\uparrow$&S$\downarrow$&H$\uparrow$&B$\uparrow$&F$\uparrow$&S$\downarrow$&H$\uparrow$&B$\uparrow$&F$\uparrow$&S$\downarrow$&H$\uparrow$&B$\uparrow$\\
   \midrule 
     Zero-CoT &\textbf{71.3}&29.1&1.6&1.2&46.9&11.0&1.8&1.1&44.6&9.5&1.7&1.0&\textbf{47.2}&24.1&1.8&1.1&37.3&13.0&2&1.0&46.4&9.9&1.8&1.1\\
     Auto-CoT &48.2&9.9&1.4&0.9&48.3&\textbf{4.5}&1.6&1.0&\textbf{46.4}&9.4&1.5&0.9&42.4&7.6&1.7&1.0&45.7&\textbf{8.8}&1.7&1.0&45.4&\textbf{8.9}&2.1&1.2\\
     APE &13.7&11.1&1.5&0.6&47.2&13.4&1.6&1.0&36.4&15.3&1.4&0.8&45.0&7.6&1.7&1.0&40.1&14.1&1.6&0.9&39.3&13.9&1.6&0.9\\
     PS &45.9&16.2&1.2&0.8&\textbf{50.9}&8.0&1.5&1.0&40.3&11.8&1.5&0.9&45.3&7.6&1.9&1.1&39.3&15.2&1.7&0.9&41.2&13.7&1.8&1.0\\
     PS+ &48.6&19.3&2.1&1.2&49.6&8.8&2.2&1.2&46.1&9.4&\textbf{2.2}&1.2&45.5&8.1&2.1&\textbf{1.2}&40.3&13.2&1.7&1.0&\textbf{47.2}&11.9&1.9&1.1\\
     \midrule
    IDADP\_GPT &49.3&\textbf{8.0}&\textbf{2.6}&\textbf{1.4}&47.9&9.4&\textbf{2.5}&\textbf{1.3}&43.7&\textbf{8.5}&\textbf{2.2}&\textbf{1.2}&46.9&\textbf{7.2}&\textbf{2.6}&\textbf{1.3}&46.1&9.2&\textbf{2.6}&\textbf{1.3}&46.5&9.0&\textbf{2.5}&\textbf{1.3}\\
     IDADP\_Gemini &45.4&9.9&2.2&1.2&46.8&9.4&2.2&1.2&43.6&9.0&2.0&1.2&46.7&7.6&2.4&1.3&\textbf{46.6}&10.4&2.6&\textbf{1.3}&40.5&11.2&2.4&1.2\\
     \bottomrule
    \end{tabular}
    }
 \end{center}
 \caption{Results of the IDADP in comparison to the baselines across six datasets for reasoning. F: Flesch Reading Ease Score; S: Standard Deviation; H: human annotation score; B: Balance of F and H; The best performances are highlighted in bold.}
 \label{tab:results_reason}
\end{table*}

\begin{table}
\scalebox{0.95}{
\begin{tabular}{lll}
   
    \textbf{\textit{Sample:}}&{\textit{Max Verstappen is such a clean driver,}}&Ironic                \\
    &{\textit{he never makes dirty moves when racing.}}&                \\
    \toprule
   \textbf{Approach}&\textbf{Reasoning }& \textbf{FRE} \\
   \midrule
   \textbf{CoT}& \textit{Max Verstappen is actually known for making}   &33.7\\
   & \textit{aggressive and sometimes reckless moves} &\\
    &  \textit{when racing, so describing him as a "clean}&\\
    &  \textit{driver “who" never makes dirty moves" is}&\\
    &  \textit{ironic.}&\\
    \midrule
    \textbf{Auto-CoT}&\textit{It is ironic because it sarcastically praises Max}  &27.5\\
    & \textit{Verstappen as a clean driver who never makes} &\\ &\textit{dirty moves when racing, when in reality,}&\\
    &\textit{Verstappen has been involved in several}&\\
    &\textit{controversial incidents on the track.}&\\
   \midrule
   \textbf{APE}&\textit{Max Verstappen is known for his aggressive}&43.6\\& \textit{driving style and has been involved in several}&\\
   & \textit{controversial incidents on the track, so the}&\\
   &\textit{statement that he is a clean driver who never}&\\
   &\textit{makes dirty moves is ironic.}&\\

   \midrule
   \textbf{PS}&\textit{It is a straightforward statement praising Max}&57.1\\
    &\textit{mVerstappen for his clean driving.}&\\
   \midrule
   \textbf{PS+}&\textit{The use of the word “clean" to describe a} &47.7\\ 
   & \textit{driver in a racing context is ironic because it} &\\&\textit{typically refers to a driver who avoids making} &\\
   &  \textit{aggressive or risky moves on the track.} &\\
   &\textit{However, the statement goes on to mention}&\\
   &\textit{that the driver “never makes dirty moves}&\\
   &\textit{when racing," which implies that the driver is}&\\
   &\textit{actually not clean in their racing tactics.}&\\
   &\textit{Therefore, the statement is ironic.}&\\
   \midrule
   \textbf{IDADP\_GPT}&\textit{Yes, the statement Max Verstappen is such a} &54.2\\& \textit{clean driver, he never makes dirty moves} &\\&\textit{when racing." is likely ironic. The irony arises} &\\&\textit{from the contrast between the literal praise of} &\\&\textit{being a "clean driver" and the implied}&\\
   &\textit{sarcasm, suggesting that Verstappen is}&\\
   &\textit{actually known for making aggressive or}&\\
   &\textit{controversial moves in races. The tone implies}&\\
   &\textit{the opposite of what the words directly state,}&\\
   &\textit{which is a hallmark of irony.}&\\   
    \bottomrule
  \end{tabular}
  }
   \caption{Flesch Reading Ease Score (FRE) for reasoning Examples Across Six Models.}\label{tab:results_reason_error}
\end{table}

 \begin{figure}[tbh]
  \centering
  \includegraphics[width=\linewidth]
  {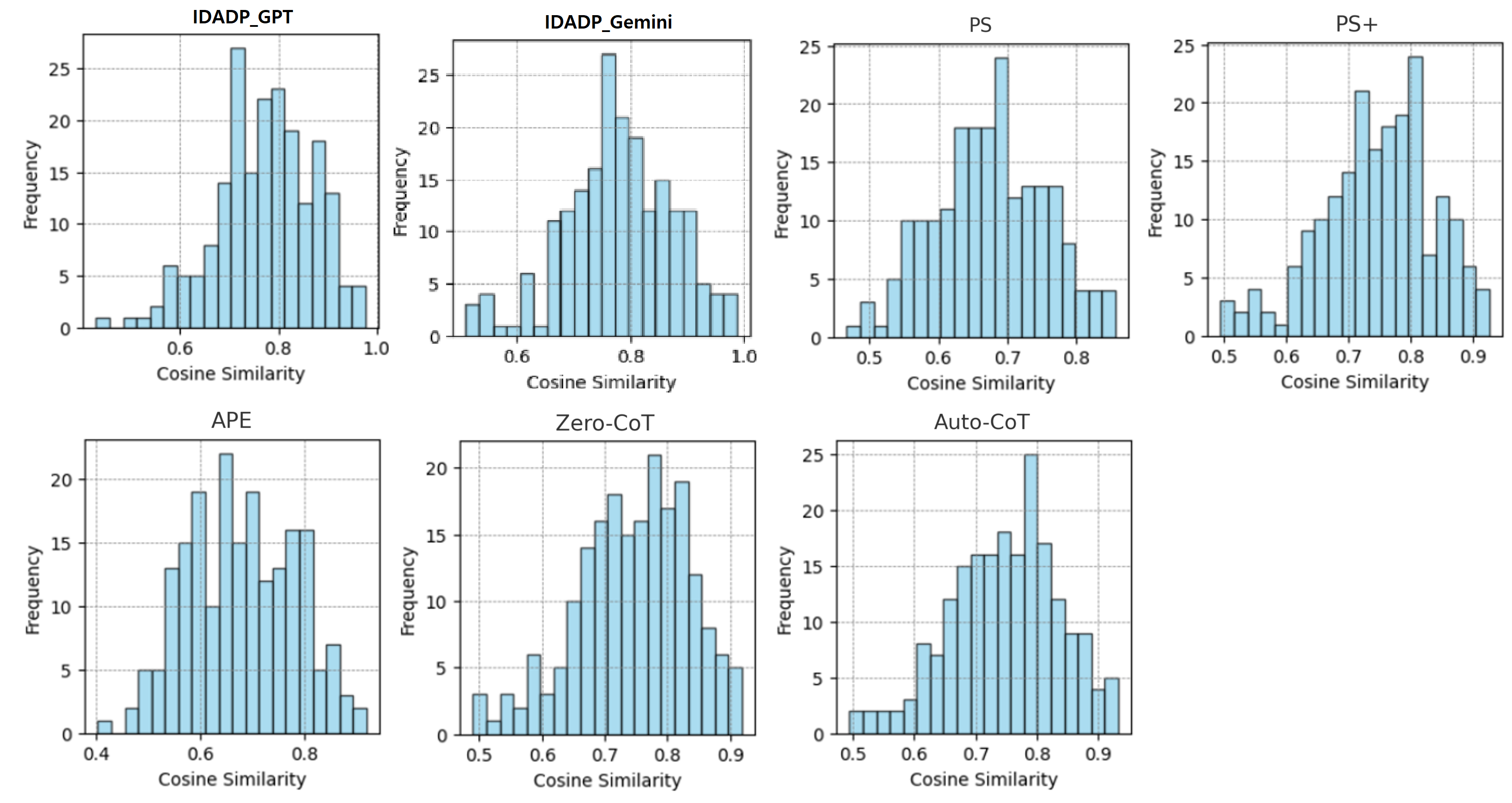}
  \caption {Frequency distribution of cosine similarity scores for irony understanding across six models.}
  \label{fig:understanding_cosin_1}
\end{figure}

\begin{figure}[tbh]
  \centering
  \includegraphics[width=\linewidth]
  {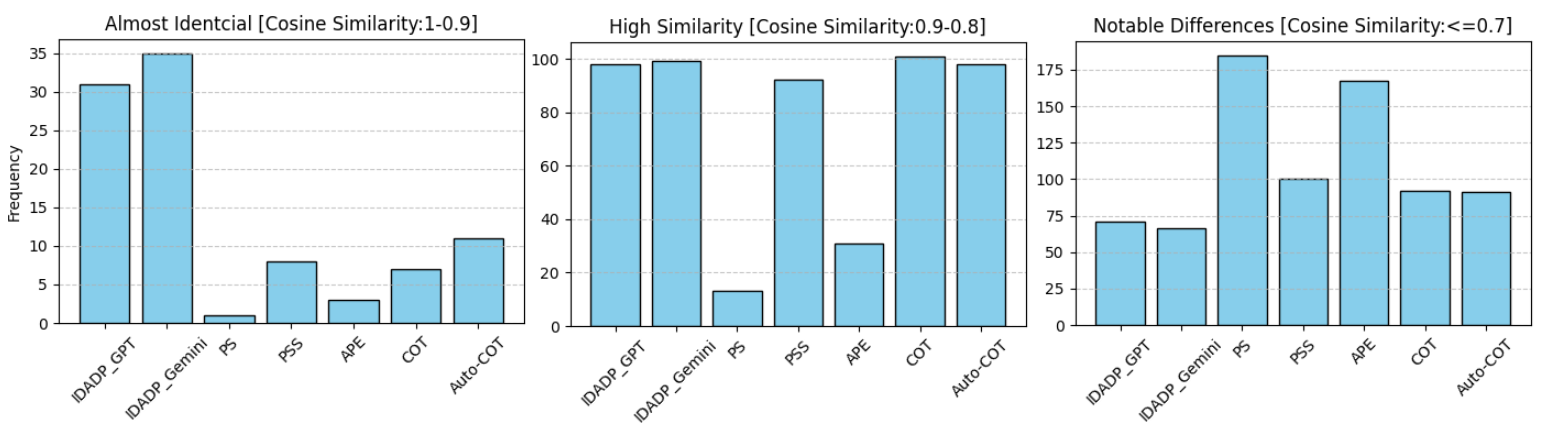}
  \caption {Frequency distribution of cosine similarity scores across three ranges for irony understanding.}
  \label{fig:Undertanding_cosion_2}
\end{figure}

\begin{figure}[tbh]
  \centering
  \includegraphics[width=\linewidth]
  {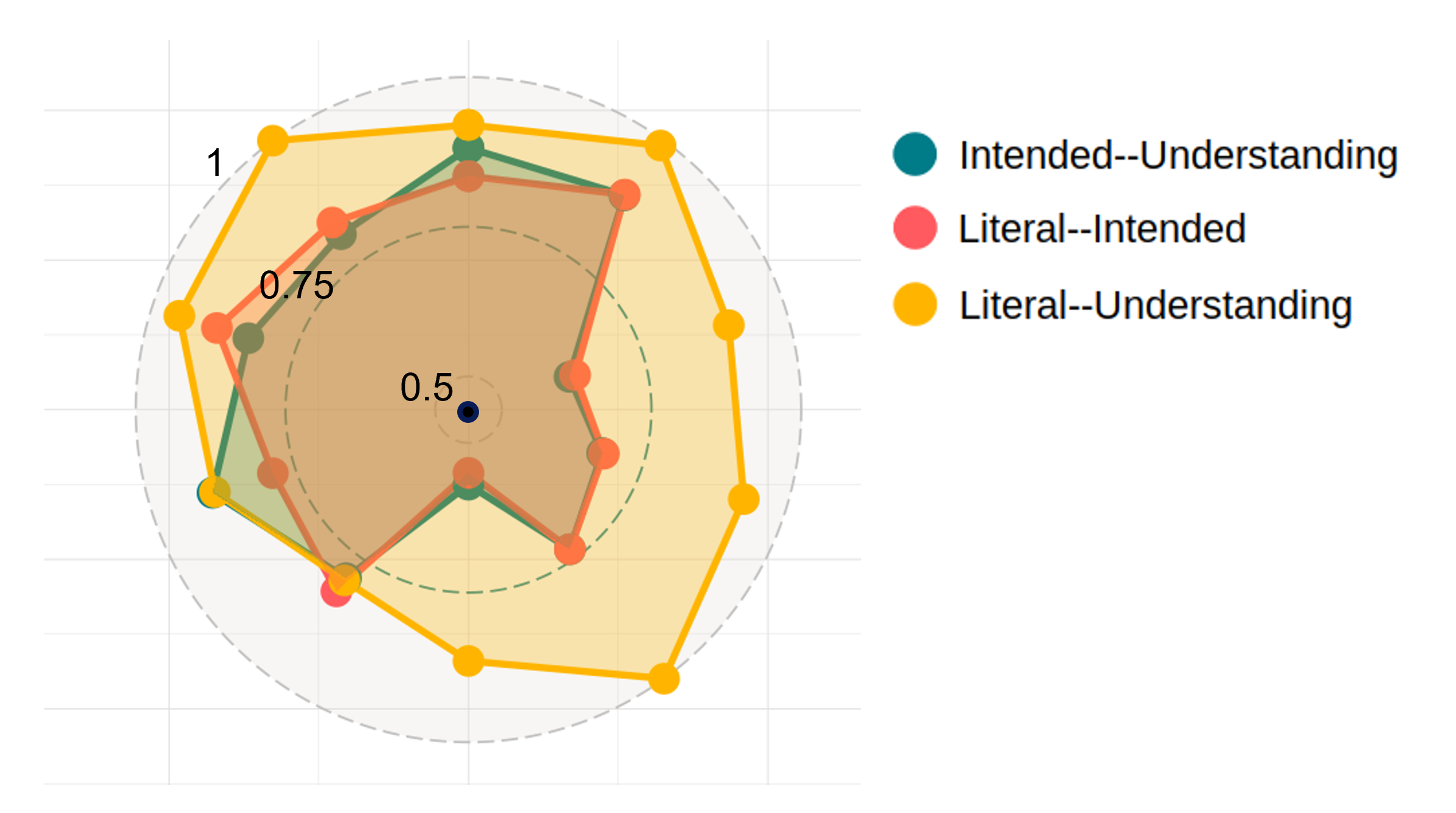}
  \caption {Radar charts display cosine similarities of literal, intended, and understanding meanings, where vertices represent similarity values with higher scores closer to the chart's outer edge.}
  \label{fig:understanding_similarity}
  \end{figure}
  
\subsubsection{Detection and Generalization}

Table \ref{tab:detection-results} presents the performance of the proposed framework (IDADP) compared to six zero-shot baseline models and three fine-tuning models across six irony detection datasets. We can observe: 1) \textbf{Approach performance with fine-tuned models.} IDADP achieves performance comparable to fine-tuned models, despite not using any training data. It shows only a small gap in overall performance and outperforms all fine-tuned models on the RQ dataset. 
\textbf{2) Significantly surpasses zero-shot GPT3.5 and Gemini1.5 based models by a large margin in F1-score.}  And consistently outperformed in precision and recall. Notably, no other zero-shot method performed well across all datasets, highlighting IDADP's strong generalization ability. 3) \textbf{Auto-CoT and PS+ exhibit the worst performance,} indicating that for complex tasks like irony detection, we cannot rely solely on general knowledge from pre-trained data or automatic learning during the task. 4) When IDADP is applied to both GPT-3.5 and Gemini 1.5, we observe significant performance improvements across both models, demonstrating the effectiveness of our framework. Furthermore, IDADP achieves comparable results on each base model, indicating that it delivers robust performance regardless of the underlying language model.  

To gain further insight into IDADP effectiveness, we conducted a comparative analysis of various GPT-based models on both ironic and non-ironic classification using precision and recall metrics. Some important phenomena we can observe from Figure \ref{fig:positive}: \textbf{1) Balance between precision and recall.} The IDADP model (black point) consistently maintains a strong balance between precision and recall across most datasets. In contrast, baseline models demonstrate greater variability, with some prioritizing precision over recall, and others exhibiting the reverse, suggesting limitations in their generalization capabilities. \textbf{2) A performance divergence is observed between irony and non-irony detection.} Models generally exhibit lower precision in irony detection but higher precision in non-irony detection, underscoring the inherent complexity of identifying irony. IDADP still requires further refinement, especially in minimizing false positives when classifying non-ironic statements. \textbf{3) Contextual variations significantly impact irony detection accuracy.} Model balance varies significantly across datasets, highlighting dataset-specific challenges. For instance, while all models perform well on the HYP dataset, they struggle on the Semeval dataset.

To diagnose IDADP weaknesses, improve performance, and ensure reliability, Table \ref{tab:misclass} lists various posts with misclassification along with their predicted labels and human evaluations. The errors can be categorized as false positives (predicted ironic when not) and false negatives (predicted non-ironic when it is). We can generally classify the three reasons the model makes wrong predictions.
\textbf{1) Contextual and Linguistic Complexity:} The model struggles with contextual nuances and the implications of certain phrases, such as "humans like," as seen in samples [1, 3, 4, 6].
\textbf{2) Model Calibration:} Many simple questions or comments often fail to be recognized by the model when expressed sarcastically, revealing underlying gaps in subtle language processing capabilities, as seen in samples [2, 4]. \textbf{3) Need for Enhanced Training:} Improving model performance may necessitate exposure to more diverse linguistic structures, focusing specifically on irony, sarcasm, and social commentary,as seen in samples [5, 6].


\subsubsection{Reasoning}
Table \ref{tab:results_reason} compares the Flesch-Kincaid readability score (F), Standard Deviation (S) of F, Human Evaluation Scores (H) and Balance (B) of F and H for different prompts approaches across several datasets. In general, IDADP demonstrates \textbf{1) A strong balance (B)} between readability (F) and human evaluation (H) across all datasets. \textbf{2) Performs well (average 2.5) in Human Evaluations (H)}, showing its outputs are well received by humans in these datasets. \textbf{3) Relatively lower standard deviations (S)} across datasets, which is a positive sign for predictability and stability in response quality. \textbf{4) Moderately high but not the highest on Flesch Reading Ease score (F)}. 5) IDADP\_GPT slightly outperforms IDADP\_Gemini on most F1 scores and other metrics, but the gap is relatively small.

Meanwhile, we notice some models achieve the highest Flesch Reading Ease scores but with very low human annotator value. For example, PS achieves a particularly high value(50.9), but with the lowest human annotator score (1.5) on the SemEval dataset. The phenomena highlight the shortcomings of the Flesch Reading Ease score: it relies solely on sentence length and word complexity to assess readability. Table \ref{tab:results_reason_error} illustrates the phenomena: while oversimplified sentences may achieve a high Flesch Reading Ease score, stripping away important nuances, details, or sophistication from the text can result in a loss of essential context, accuracy, or depth.


\subsubsection{Understanding}

Figure \ref{fig:understanding_cosin_1} plots
the frequency of cosine similarity scores for the corresponding modes, which quantifies the alignment between the generated sentence meaning and the actual meaning.
 Higher scores indicate model rephrases ironic sentences in a non-ironic manner while preserving the intended meaning. 
IDADP  consistently exhibits high cosine similarity scores, ranging from 0.7 to 0.9. Reflecting its effectiveness in capturing the intended meaning. In contrast, lower scores indicate a failure to preserve the intended meaning or a shift towards unintended interpretations. APE exhibits a broader range of scores, from 0.4 to 0.9, peaking around 0.7, which suggests a lack of understanding of irony.  While PS and PS+ perform moderately well, they lack consistency like IDADP. Auto-CoT and CoT show tight distributions between 0.7 and 0.9 but still fall behind IDADP. 

To further distinguish the performance of each model in irony understanding, Figure \ref{fig:Undertanding_cosion_2} presents the frequency distribution of cosine
similarity scores across three categories. IDADP demonstrably achieves a high concentration of sentences in the "Almost Identical " and a minimal number in the "Notable Differences". Highlighting its superior performance.

To gain a deeper understanding of the model's comprehension of irony, Figure \ref{fig:understanding_similarity} presents the alignment (cosine similarity) between literal meanings (original sentences), intended meanings (author-provided), and the understated meanings generated by IDADP\_GPT. The consistently higher cosine similarity scores observed for the Literal-Understanding (orange line) reflect that the model's comprehension heavily relies on the literal interpretation of the text. The fluctuating scores in the Literal-Intended (red line) and Intended-Understanding (blue line) comparisons indicate challenges in accurately discerning irony, likely due to the absence of explicit ironic markers and insufficient contextual information, as exemplified in Table \ref{tab:misclass_understanding}. 
\begin{table}[h]
\centering
\scalebox{0.90}{
\centering
  
  \begin{tabular}{lll}
    \toprule
    Posts & Intended&Understanding                 \\
        \midrule
   I see that your team   &I'm sorry that your team&I see that your team\\
    played well today!  & didn’t win yesterday.& played poorly today.\\
 \midrule
    
 I think my husband has&I know you're busy but&My husband seems to\\
lost the ability to find&can I have a cuppa &have trouble locating \\
 the kitchen.&please?&the kitchen.\\
    \midrule 
 Dry your eyes mate,&The way people wear &Dry your eyes mate, \\
  they're only trousers.&their trousers does not &they're only trousers.\\
  &matter to most people&\\
   &so perhaps it would be&\\
    &a better use of your time &\\
     &to focus on yourself.&\\
    \bottomrule
  \end{tabular}
}
\caption{Three samples of misunderstanding Intent.}
 \label{tab:misclass_understanding}
\end{table}

\subsection {Ablation Study}
We conduct the ablation study on the detection task, as it allows for a clearer and more objective evaluation of each component’s contribution, minimising the ambiguity that may arise in more subjective tasks such as irony understanding or reasoning. To assess the contributions of \textbf{(1) the voting mechanism} and \textbf{(2) knowledge integration} in the IDADP framework, we systematically remove each component and evaluate the resulting impact on performance.

Table \ref{tab:ablation_compact} presents the results of the ablation study on the irony detection task across six datasets. The IDADP framework achieves the highest performance on all datasets, demonstrating the effectiveness of combining multiple prompts through the voting mechanism. When evaluating individual prompts without voting (P1, P2, P3), performance generally decreases compared to the full IDADP framework, highlighting the benefit of aggregating multiple responses. Furthermore, removing knowledge integration (NI) results in a further decline in performance, indicating that incorporating extracted knowledge is crucial for achieving optimal results. These findings confirm that both the voting mechanism and knowledge integration are important contributors to the overall effectiveness and robustness of the IDADP framework.

\begin{table}[!ht]
\centering

\begin{tabular}{lccccc}
\hline
\textbf{Dataset} & \textbf{IDADP\_GPT} & \textbf{(P1)} & \textbf{(P2)} & \textbf{(P3)} & \textbf{NI} \\
\hline
iSarcasm  & 0.67 & 0.60 & 0.59& 0.56 & 0.60 \\
SemEval & 0.65 & 0.66 & 0.64 & 0.62 & 0.59 \\
Gen & 0.74 & 0.78 & 0.70 & 0.70 & 0.70 \\
RQ & 0.82 & 0.70 & 0.73 & 0.73 & 0.65 \\
HYP & 0.76 & 0.74 & 0.74 & 0.77 & 0.69 \\
Reddit & 0.64 & 0.59 & 0.62 & 0.59 & 0.43 \\
\hline
\end{tabular}

\caption{Ablation study results(F1) for irony detection across six benchmark datasets. “IDADP\_GPT”: refers to the full model. “P1”, “P2”, and “P3” correspond to results from individual prompts without voting. “NI”: the setting without knowledge integration. }
\label{tab:ablation_compact}
\end{table}

\section{Discussion and Conclusion}

Irony detection is a complex linguistic task that involves recognising when the intended meaning of a statement differs from its literal interpretation, especially in zero-shot learning scenarios where models lack pre-training on ironic examples. Our research leverages LLMs’ zero-shot capabilities to detect irony across various social media platforms without relying on existing datasets. We introduce the IDADP framework, enhancing interactions through effective prompt engineering, improved context awareness, and features that address irony contradictions. Our refined evaluation framework incorporates linguistic and contextual analysis, enabling a more comprehensive assessment of performance, reliability, and effectiveness in irony detection, reasoning, and understanding tasks. 
Below, We will discuss key observations, how our findings address identified limitations, and the ongoing challenges.

\textbf{Generalisation :} To address generalisation challenges, we
leveraged GPT and Gemini's zero-shot learning and prompt engineering to improve adaptability to unseen data, reducing overfitting. Task-specific prompts improved handling of linguistic complexities, though misclassification persists due to three main
issues: First, the model struggles with nuanced meanings due
to contextual and linguistic complexity. Second, model calibration remains a limitation, particularly in recognising subtle
cues like irony. Lastly, the model’s accuracy could benefit from
exposure to a broader range of linguistic patterns, especially
those involving irony, sarcasm, and social commentary.

\textbf{Reasoning:}
In our experiments, we found that integrating
prompt engineering with multi-step reasoning processes and irony-focused knowledge significantly enhanced LLMs’
ability to generate coherent, human-readable reasoning. However, there was a significant difference in performance when
the model encountered unfamiliar contexts, highlighting the
importance of providing rich contextual information to enhance the model’s focus.

\textbf{Understanding:} 
We analysed GPT and Gemini capability to convert 200 ironic texts into their non-ironic equivalents, and we
found that these model demonstrated a notable proficiency in
understanding irony. It was able to retain the intended message
by relying on literal interpretations. However, this approach
is limited in scenarios where explicit markers or contextual
clues are absent. While the task underscored LLMs’ ability
to disentangle complex linguistic features, it simultaneously
revealed its limitations in comprehensively grasping irony
without clear indicators. For example, irony's understanding relies entirely on common sense.

\textbf{Future directions:} 
Enhancing LLMS’ zero-shot capabilities for irony detection, reasoning, and comprehension offers promising research paths. Key areas include improving contextual awareness for irony detection, as it relies on subtle, conversation-embedded cues; Context-aware attention mechanisms or expanded context windows could help capture these nuances. Hybrid symbolic-neural methods may support complex reasoning by combining logical structuring with neural flexibility. Addressing implicit knowledge gaps is also crucial, as irony often requires shared cultural understanding; Multi-task transfer learning across satire, hypothetical reasoning, and idiom-related tasks could strengthen this. Finally, integrating multimodal data, such as images and videos, could enhance irony detection by providing richer contextual understanding, paving the way for a more nuanced LLM.

\bibliographystyle{IEEEtran}
\bibliography{Irony.bib}

\begin{thebibliography}{10}
\providecommand{\url}[1]{#1}
\csname url@samestyle\endcsname
\providecommand{\newblock}{\relax}
\providecommand{\bibinfo}[2]{#2}
\providecommand{\BIBentrySTDinterwordspacing}{\spaceskip=0pt\relax}
\providecommand{\BIBentryALTinterwordstretchfactor}{4}
\providecommand{\BIBentryALTinterwordspacing}{\spaceskip=\fontdimen2\font plus
\BIBentryALTinterwordstretchfactor\fontdimen3\font minus \fontdimen4\font\relax}
\providecommand{\BIBforeignlanguage}[2]{{%
\expandafter\ifx\csname l@#1\endcsname\relax
\typeout{** WARNING: IEEEtran.bst: No hyphenation pattern has been}%
\typeout{** loaded for the language `#1'. Using the pattern for}%
\typeout{** the default language instead.}%
\else
\language=\csname l@#1\endcsname
\fi
#2}}
\providecommand{\BIBdecl}{\relax}
\BIBdecl

\bibitem{booth1974rhetoric}
W.~C. Booth, \emph{A rhetoric of irony}.\hskip 1em plus 0.5em minus 0.4em\relax University of Chicago Press, 1974.

\bibitem{lear2011case}
J.~Lear, \emph{A case for irony}.\hskip 1em plus 0.5em minus 0.4em\relax Harvard University Press, 2011.

\bibitem{joshi2017automatic}
A.~Joshi, P.~Bhattacharyya, and M.~J. Carman, ``Automatic sarcasm detection: A survey,'' \emph{ACM Computing Surveys (CSUR)}, vol.~50, no.~5, pp. 1--22, 2017.

\bibitem{lucariello1994situational}
J.~Lucariello, ``Situational irony: A concept of events gone awry.'' \emph{Journal of Experimental Psychology: General}, vol. 123, no.~2, p. 129, 1994.

\bibitem{reyes2012humor}
A.~Reyes, P.~Rosso, and D.~Buscaldi, ``From humor recognition to irony detection: The figurative language of social media,'' \emph{Data \& Knowledge Engineering}, vol.~74, pp. 1--12, 2012.

\bibitem{potamias2020transformer}
R.~A. Potamias, G.~Siolas, and A.-G. Stafylopatis, ``A transformer-based approach to irony and sarcasm detection,'' \emph{Neural Computing and Applications}, vol.~32, no.~23, pp. 17\,309--17\,320, 2020.

\bibitem{jang2024generalizable}
H.~Jang and D.~Frassinelli, ``Generalizable sarcasm detection is just around the corner, of course!'' \emph{arXiv preprint arXiv:2404.06357}, 2024.

\bibitem{van2018we}
C.~Van~Hee, E.~Lefever, and V.~Hoste, ``We usually don’t like going to the dentist: Using common sense to detect irony on twitter,'' \emph{Computational Linguistics}, vol.~44, no.~4, pp. 793--832, 2018.

\bibitem{farha2022semeval}
I.~A. Farha, S.~Oprea, S.~Wilson, and W.~Magdy, ``Semeval-2022 task 6: isarcasmeval, intended sarcasm detection in english and arabic,'' in \emph{The 16th International Workshop on Semantic Evaluation 2022}.\hskip 1em plus 0.5em minus 0.4em\relax Association for Computational Linguistics, 2022, pp. 802--814.

\bibitem{van2018semeval}
C.~Van~Hee, E.~Lefever, and V.~Hoste, ``Semeval-2018 task 3: Irony detection in english tweets,'' in \emph{Proceedings of the 12th international workshop on semantic evaluation}, 2018, pp. 39--50.

\bibitem{yang2024harnessing}
J.~Yang, H.~Jin, R.~Tang, X.~Han, Q.~Feng, H.~Jiang, S.~Zhong, B.~Yin, and X.~Hu, ``Harnessing the power of llms in practice: A survey on chatgpt and beyond,'' \emph{ACM Transactions on Knowledge Discovery from Data}, vol.~18, no.~6, pp. 1--32, 2024.

\bibitem{brown2020language}
T.~B. Brown, ``Language models are few-shot learners,'' \emph{arXiv preprint arXiv:2005.14165}, 2020.

\bibitem{cheng2024chainlm}
X.~Cheng, J.~Li, W.~X. Zhao, and J.-R. Wen, ``Chainlm: Empowering large language models with improved chain-of-thought prompting,'' \emph{arXiv preprint arXiv:2403.14312}, 2024.

\bibitem{wei2022chain}
J.~Wei, X.~Wang, D.~Schuurmans, M.~Bosma, F.~Xia, E.~Chi, Q.~V. Le, D.~Zhou \emph{et~al.}, ``Chain-of-thought prompting elicits reasoning in large language models,'' \emph{Advances in neural information processing systems}, vol.~35, pp. 24\,824--24\,837, 2022.

\bibitem{hariri2023unlocking}
W.~Hariri, ``Unlocking the potential of chatgpt: A comprehensive exploration of its applications, advantages, limitations, and future directions in natural language processing,'' \emph{arXiv preprint arXiv:2304.02017}, 2023.

\bibitem{ouyang2022training}
L.~Ouyang, J.~Wu, X.~Jiang, D.~Almeida, C.~Wainwright, P.~Mishkin, C.~Zhang, S.~Agarwal, K.~Slama, A.~Ray \emph{et~al.}, ``Training language models to follow instructions with human feedback,'' \emph{Advances in neural information processing systems}, vol.~35, pp. 27\,730--27\,744, 2022.

\bibitem{ray2023chatgpt}
P.~P. Ray, ``Chatgpt: A comprehensive review on background, applications, key challenges, bias, ethics, limitations and future scope,'' \emph{Internet of Things and Cyber-Physical Systems}, vol.~3, pp. 121--154, 2023.

\bibitem{white2023prompt}
J.~White, Q.~Fu, S.~Hays, M.~Sandborn, C.~Olea, H.~Gilbert, A.~Elnashar, J.~Spencer-Smith, and D.~C. Schmidt, ``A prompt pattern catalog to enhance prompt engineering with chatgpt,'' \emph{arXiv preprint arXiv:2302.11382}, 2023.

\bibitem{azaria2024chatgpt}
A.~Azaria, R.~Azoulay, and S.~Reches, ``Chatgpt is a remarkable tool—for experts,'' \emph{Data Intelligence}, vol.~6, no.~1, pp. 240--296, 2024.

\bibitem{bharti2015parsing}
S.~K. Bharti, K.~S. Babu, and S.~K. Jena, ``Parsing-based sarcasm sentiment recognition in twitter data,'' in \emph{Proceedings of the 2015 IEEE/ACM International Conference on Advances in Social Networks Analysis and Mining 2015}, 2015, pp. 1373--1380.

\bibitem{riloff2013sarcasm}
E.~Riloff, A.~Qadir, P.~Surve, L.~De~Silva, N.~Gilbert, and R.~Huang, ``Sarcasm as contrast between a positive sentiment and negative situation,'' in \emph{Proceedings of the 2013 conference on empirical methods in natural language processing}, 2013, pp. 704--714.

\bibitem{mladenovic2017using}
M.~Mladenovi{\'c}, C.~Krstev, J.~Mitrovi{\'c}, and R.~Stankovi{\'c}, ``Using lexical resources for irony and sarcasm classification,'' in \emph{Proceedings of the 8th Balkan Conference in Informatics}, 2017, pp. 1--8.

\bibitem{gonzalez2011identifying}
R.~Gonz{\'a}lez-Ib{\'a}nez, S.~Muresan, and N.~Wacholder, ``Identifying sarcasm in twitter: a closer look,'' in \emph{Proceedings of the 49th annual meeting of the association for computational linguistics: human language technologies}, 2011, pp. 581--586.

\bibitem{maynard2014cares}
D.~G. Maynard and M.~A. Greenwood, ``Who cares about sarcastic tweets? investigating the impact of sarcasm on sentiment analysis,'' in \emph{Lrec 2014 proceedings}.\hskip 1em plus 0.5em minus 0.4em\relax ELRA, 2014.

\bibitem{dimovska2018sarcasm}
J.~Dimovska, M.~Angelovska, D.~Gjorgjevikj, and G.~Madjarov, ``Sarcasm and irony detection in english tweets,'' in \emph{ICT Innovations 2018. Engineering and Life Sciences: 10th International Conference, ICT Innovations 2018, Ohrid, Macedonia, September 17--19, 2018, Proceedings 10}.\hskip 1em plus 0.5em minus 0.4em\relax Springer, 2018, pp. 120--131.

\bibitem{fersini2015detecting}
E.~Fersini, F.~A. Pozzi, and E.~Messina, ``Detecting irony and sarcasm in microblogs: The role of expressive signals and ensemble classifiers,'' in \emph{2015 IEEE international conference on data science and advanced analytics (DSAA)}.\hskip 1em plus 0.5em minus 0.4em\relax IEEE, 2015, pp. 1--8.

\bibitem{hernandez2015applying}
I.~Hern{\'a}ndez-Far{\'\i}as, J.-M. Bened{\'\i}, and P.~Rosso, ``Applying basic features from sentiment analysis for automatic irony detection,'' in \emph{Pattern Recognition and Image Analysis: 7th Iberian Conference, IbPRIA 2015, Santiago de Compostela, Spain, June 17-19, 2015, Proceedings 7}.\hskip 1em plus 0.5em minus 0.4em\relax Springer, 2015, pp. 337--344.

\bibitem{karoui2015towards}
J.~Karoui, F.~Benamara, V.~Moriceau, N.~Aussenac-Gilles, and L.~H. Belguith, ``Towards a contextual pragmatic model to detect irony in tweets,'' in \emph{53rd Annual Meeting of the Association for Computational Linguistics (ACL 2015)}, vol.~2.\hskip 1em plus 0.5em minus 0.4em\relax ACL: Association for Computational Linguistics, 2015, pp. 644--650.

\bibitem{kumar2023empirical}
A.~Kumar and G.~Garg, ``Empirical study of shallow and deep learning models for sarcasm detection using context in benchmark datasets,'' \emph{Journal of ambient intelligence and humanized computing}, vol.~14, no.~5, pp. 5327--5342, 2023.

\bibitem{huang2017irony}
Y.-H. Huang, H.-H. Huang, and H.-H. Chen, ``Irony detection with attentive recurrent neural networks,'' in \emph{Advances in Information Retrieval: 39th European Conference on IR Research, ECIR 2017, Aberdeen, UK, April 8-13, 2017, Proceedings 39}.\hskip 1em plus 0.5em minus 0.4em\relax Springer, 2017, pp. 534--540.

\bibitem{ahuja2022transformer}
R.~Ahuja and S.~C. Sharma, ``Transformer-based word embedding with cnn model to detect sarcasm and irony,'' \emph{Arabian Journal for Science and Engineering}, vol.~47, no.~8, pp. 9379--9392, 2022.

\bibitem{olaniyan2023utilizing}
D.~Olaniyan, R.~O. Ogundokun, O.~P. Bernard, J.~Olaniyan, R.~Maskeli{\=u}nas, and H.~B. Akande, ``Utilizing an attention-based lstm model for detecting sarcasm and irony in social media,'' \emph{Computers}, vol.~12, no.~11, p. 231, 2023.

\bibitem{belal2023leveraging}
M.~Belal, J.~She, and S.~Wong, ``Leveraging chatgpt as text annotation tool for sentiment analysis,'' \emph{arXiv preprint arXiv:2306.17177}, 2023.

\bibitem{gole2023sarcasm}
M.~Gole, W.-P. Nwadiugwu, and A.~Miranskyy, ``On sarcasm detection with openai gpt-based models,'' \emph{arXiv preprint arXiv:2312.04642}, 2023.

\bibitem{aytekin2023generative}
M.~U. Aytekin and O.~A. Erdem, ``Generative pre-trained transformer (gpt) models for irony detection and classification,'' in \emph{2023 4th International Informatics and Software Engineering Conference (IISEC)}.\hskip 1em plus 0.5em minus 0.4em\relax IEEE, 2023, pp. 1--8.

\bibitem{radford2018improving}
A.~Radford, ``Improving language understanding by generative pre-training,'' 2018.

\bibitem{radford2019language}
A.~Radford, J.~Wu, R.~Child, D.~Luan, D.~Amodei, I.~Sutskever \emph{et~al.}, ``Language models are unsupervised multitask learners,'' \emph{OpenAI blog}, vol.~1, no.~8, p.~9, 2019.

\bibitem{kojima2022large}
T.~Kojima, S.~S. Gu, M.~Reid, Y.~Matsuo, and Y.~Iwasawa, ``Large language models are zero-shot reasoners,'' \emph{Advances in neural information processing systems}, vol.~35, pp. 22\,199--22\,213, 2022.

\bibitem{wang2022self}
X.~Wang, J.~Wei, D.~Schuurmans, Q.~Le, E.~Chi, S.~Narang, A.~Chowdhery, and D.~Zhou, ``Self-consistency improves chain of thought reasoning in language models,'' \emph{arXiv preprint arXiv:2203.11171}, 2022.

\bibitem{zhang2022automatic}
Z.~Zhang, A.~Zhang, M.~Li, and A.~Smola, ``Automatic chain of thought prompting in large language models,'' \emph{arXiv preprint arXiv:2210.03493}, 2022.

\bibitem{yao2024tree}
S.~Yao, D.~Yu, J.~Zhao, I.~Shafran, T.~Griffiths, Y.~Cao, and K.~Narasimhan, ``Tree of thoughts: Deliberate problem solving with large language models,'' \emph{Advances in Neural Information Processing Systems}, vol.~36, 2024.

\bibitem{long2023large}
J.~Long, ``Large language model guided tree-of-thought,'' \emph{arXiv preprint arXiv:2305.08291}, 2023.

\bibitem{hulbert2023using}
D.~Hulbert, ``Using tree-of-thought prompting to boost chatgpt’s reasoning,'' 2023.

\bibitem{liu2021generated}
J.~Liu, A.~Liu, X.~Lu, S.~Welleck, P.~West, R.~L. Bras, Y.~Choi, and H.~Hajishirzi, ``Generated knowledge prompting for commonsense reasoning,'' \emph{arXiv preprint arXiv:2110.08387}, 2021.

\bibitem{yang2023large}
C.~Yang, X.~Wang, Y.~Lu, H.~Liu, Q.~V. Le, D.~Zhou, and X.~Chen, ``Large language models as optimizers,'' 2023.

\bibitem{fernando2023promptbreeder}
C.~Fernando, D.~Banarse, H.~Michalewski, S.~Osindero, and T.~Rockt{\"a}schel, ``Promptbreeder: Self-referential self-improvement via prompt evolution,'' \emph{arXiv preprint arXiv:2309.16797}, 2023.

\bibitem{zhou2022large}
Y.~Zhou, A.~I. Muresanu, Z.~Han, K.~Paster, S.~Pitis, H.~Chan, and J.~Ba, ``Large language models are human-level prompt engineers,'' \emph{arXiv preprint arXiv:2211.01910}, 2022.

\bibitem{kong2024prewrite}
W.~Kong, S.~A. Hombaiah, M.~Zhang, Q.~Mei, and M.~Bendersky, ``Prewrite: Prompt rewriting with reinforcement learning,'' \emph{arXiv preprint arXiv:2401.08189}, 2024.

\bibitem{wang2023plan}
L.~Wang, W.~Xu, Y.~Lan, Z.~Hu, Y.~Lan, R.~K.-W. Lee, and E.-P. Lim, ``Plan-and-solve prompting: Improving zero-shot chain-of-thought reasoning by large language models,'' \emph{arXiv preprint arXiv:2305.04091}, 2023.

\bibitem{wei2021finetuned}
J.~Wei, M.~Bosma, V.~Y. Zhao, K.~Guu, A.~W. Yu, B.~Lester, N.~Du, A.~M. Dai, and Q.~V. Le, ``Finetuned language models are zero-shot learners,'' \emph{arXiv preprint arXiv:2109.01652}, 2021.

\bibitem{sanh2021multitask}
V.~Sanh, A.~Webson, C.~Raffel, S.~H. Bach, L.~Sutawika, Z.~Alyafeai, A.~Chaffin, A.~Stiegler, T.~L. Scao, A.~Raja \emph{et~al.}, ``Multitask prompted training enables zero-shot task generalization,'' \emph{arXiv preprint arXiv:2110.08207}, 2021.

\bibitem{oraby2017creating}
S.~Oraby, V.~Harrison, L.~Reed, E.~Hernandez, E.~Riloff, and M.~Walker, ``Creating and characterizing a diverse corpus of sarcasm in dialogue,'' \emph{arXiv preprint arXiv:1709.05404}, 2017.

\bibitem{wallace2014humans}
B.~C. Wallace, L.~Kertz, E.~Charniak \emph{et~al.}, ``Humans require context to infer ironic intent (so computers probably do, too),'' in \emph{Proceedings of the 52nd Annual Meeting of the Association for Computational Linguistics (Volume 2: Short Papers)}, 2014, pp. 512--516.

\bibitem{devlin2018bert}
J.~Devlin, ``Bert: Pre-training of deep bidirectional transformers for language understanding,'' \emph{arXiv preprint arXiv:1810.04805}, 2018.

\bibitem{liu2019roberta}
Y.~Liu, ``Roberta: A robustly optimized bert pretraining approach,'' \emph{arXiv preprint arXiv:1907.11692}, 2019.

\bibitem{song2020mpnet}
K.~Song, X.~Tan, T.~Qin, J.~Lu, and T.-Y. Liu, ``Mpnet: Masked and permuted pre-training for language understanding,'' \emph{Advances in neural information processing systems}, vol.~33, pp. 16\,857--16\,867, 2020.

\bibitem{zeng2022survey}
Q.~Zeng and A.-R. Li, ``A survey in automatic irony processing: Linguistic, cognitive, and multi-x perspectives,'' \emph{arXiv preprint arXiv:2209.04712}, 2022.

\bibitem{shalev2014understanding}
S.~Shalev-Shwartz and S.~Ben-David, \emph{Understanding machine learning: From theory to algorithms}.\hskip 1em plus 0.5em minus 0.4em\relax Cambridge university press, 2014.

\bibitem{duan2020machine}
N.~Duan, D.~Tang, and M.~Zhou, ``Machine reasoning: Technology, dilemma and future,'' in \emph{Proceedings of the 2020 Conference on Empirical Methods in Natural Language Processing: Tutorial Abstracts}, 2020, pp. 1--6.

\bibitem{bruntsch2017studying}
R.~Bruntsch and W.~Ruch, ``Studying irony detection beyond ironic criticism: Let's include ironic praise,'' \emph{Frontiers in Psychology}, vol.~8, p. 606, 2017.

\bibitem{zhang2023meta}
Y.~Zhang, ``Meta prompting for agi systems,'' \emph{arXiv preprint arXiv:2311.11482}, 2023.

\bibitem{wang2024plan}
L.~Wang, W.~Xu, Y.~Lan, Z.~Hu, Y.~Lan, R.~Lee, and E.~Lim, ``Plan-and-solve prompting: improving zero-shot chain-of-thought reasoning by large language models (2023),'' \emph{arXiv preprint arXiv:2305.04091}, 2024.

\bibitem{farr1951simplification}
J.~N. Farr, J.~J. Jenkins, and D.~G. Paterson, ``Simplification of flesch reading ease formula.'' \emph{Journal of applied psychology}, vol.~35, no.~5, p. 333, 1951.

\end{thebibliography}

\end{document}